# Reinforcement Learning by Value-Gradients

**Michael Fairbank**                                       MICHAEL.FAIRBANK 'AT' VIRGIN.NET

*c/o Gatsby Computational Neuroscience Unit,*
*Alexandra House,*
*17 Queen Square,*
*LONDON, UK*

## Abstract

The concept of the value-gradient is introduced and developed in the context of reinforcement learning, for deterministic episodic control problems that use a function approximator and have a continuous state space. It is shown that by learning the value-gradients, instead of just the values themselves, exploration or stochastic behaviour is no longer needed to find locally optimal trajectories. This is the main motivation for using value-gradients, and it is argued that learning the value-gradients is the actual objective of any value-function learning algorithm for control problems. It is also argued that learning value-gradients is significantly more efficient than learning just the values, and this argument is supported in experiments by efficiency gains of several orders of magnitude, in several problem domains.

Once value-gradients are introduced into learning, several analyses become possible. For example, a surprising equivalence between a value-gradient learning algorithm and a policy-gradient learning algorithm is proven, and this provides a robust convergence proof for control problems using a value function with a general function approximator. Also, the issue of whether to include 'residual gradient' terms into the weight update equations is addressed. Finally, an analysis is made of actor-critic architectures, which finds strong similarities to back-propagation through time, and gives simplifications and convergence proofs to certain actor-critic architectures, but while making those actor-critic architectures redundant.

Unfortunately, by proving equivalence to policy-gradient learning, finding new divergence examples even in the absence of bootstrapping, and proving the redundancy of residual-gradients and actor-critic architectures in some circumstances, this paper does somewhat discredit the usefulness of using a value-function.

## 1. Introduction

Reinforcement learning (RL) algorithms frequently make use of a value function (Sutton and Barto, 1998). On problem domains where the state space is large and continuous, the value function needs to be represented by a function approximator. In this paper, analysis is restricted to episodic control problems of this kind, with a known differentiable deterministic model.

As Sutton and Barto (1998) stated: "The central role of value estimation is arguably the most important thing we have learned about reinforcement learning over the last few decades". However the use of a value function introduces some major difficulties when combined with a function approximator, concerning the lack of convergence guarantees to learning. This problem has led to a major alternative RL approach which works without a value function at all, i.e. policy-gradient learning (PGL) algorithms (Williams, 1992; Baxter and Bartlett, 2000; Werbos, 1990), which do have the desired convergence guarantees. In this paper, a surprising equivalence between these two seemingly different approaches is shown, and this provides the basis for convergence guarantees to variants of value function learning algorithms.

It is the central thesis of this paper that for value function methods, it is not the values themselves that are important, but in fact the value-gradients (defined to be the gradient of the value function with respect to the state vector). We distinguish between methods that aim to learn a value function by explicitly updating value-gradients from those that don't by referring to them as *value-gradient learning* (VGL) and *value-learning* (VL), respectively. The necessity of exploration to VL methods is demonstrated in section 1.3, which becomes very apparent in our problem domains where all functions are deterministic. We call the level of exploration that searches immediately neighbouring trajectories as *local exploration*. This requirement for local exploration is not necessary with VGL methods, since the value-gradient automatically provides awareness of any superior neighbouring trajectories. This is shown for a specific example in Section 1.3, and proven in the general case in Appendix A. It is then argued that VGL methods are an idealised form of VL methods, are easier to analyse, and are more efficient (sections 1.4 and 1.5).

The VGL themselves are stated at the start of Section 2. One of these algorithms (Section 2.1) is proven to be equivalent to PGL. This is used as the basis for a VGL algorithm in a continuous-time formulation with convergence guarantees (Section 2.2). It also produces a tentative theoretical justification for the commonly used TD($\lambda$) weight-update, which from the author's point of view has always been a puzzling issue.

The residual-gradients algorithm for VGL is given and analysed in Section 2.3, and new reasons are given for the ineffectiveness of residual-gradients in deterministic environments, both with VGL and VL. In Section 3, actor-critic architectures are defined for VGL, and it is shown that the value-gradient analysis provides simplifications to certain actor-critic architectures. This allows new convergence proofs, but at the expense of making the actor-critic architecture redundant.

In Section 4 experimental details are provided that justify the optimality claims, and the efficiency claims (by several orders of magnitude). The problems we use include the Toy Problem, defined in Section 1.1.1, which is simple enough to be able to analyse in detail, and challenging enough to cause difficulties for VL. Examples showing diverging weights are given for all VL algorithms and some VGL algorithms in Section 4.3. Also a Lunar-Lander neural-network experiment is included, which is a larger scale neural network problem that seems to defeat VL. Finally, Section 5 gives conclusions and a discussion on VGL, highlighting the contributions of this paper.

Value-gradients have already appeared in various forms in the literature. Dayan and Singh (1996) argue the importance of value-gradients over the values themselves, which is the central thesis of this paper.

The target value-gradient we define is closely related to the "adjoint vector" that appears in Pontryagin's maximum principle, as discussed further in Appendix A.

The equations of Back-Propagation Through Time (BPTT, Werbos (1990)) and Differential Dynamic Programming (Jacobson and Mayne, 1970) implicitly contain references to the target value-gradient, both with $\lambda = 1$ ($\lambda$ is the bootstrapping parameter, defined in Section 1.1).

In continuous-time problem domains, Doya (2000) uses the value-gradient explicitly in the greedy policy, and Baird (1994) defines an "advantage" function that implicitly references the value-gradient. Both of these are discussed further in Section 2.2. A value-gradient appears in the Hamilton-Jacobi-Bellman Equation which is an optimality condition for continuous-time value-functions; although here only its component parallel to the trajectory is used, and this component is not useful in obviating the need for local exploration.

However, the most similar work on value-gradients is in a family of algorithms (Werbos, 1998; White and Sofge, 1992, ch.13) referred to as Dual Heuristic Programming (DHP). These are full VGL methods, but are based on actor-critic architectures specifically with $\lambda = 0$, and are more focussed towards unknown stochastic models.

## 1.1 Reinforcement Learning Problem Notation and Definitions

*State Space*, $S$, is a subset of $\Re^n$. Each state in the state space is denoted by a column vector $\vec{x}$. A *trajectory* is a list of states $\{\vec{x}_0, \vec{x}_1, \ldots, \vec{x}_F\}$ through state algorithm space starting at a given point $\vec{x}_0$. The trajectory is parametrised by real actions $a_t$ for time steps $t$ according to a model. The *model* is comprised of two known smooth deterministic functions $f(\vec{x}, a)$ and $r(\vec{x}, a)$. The first model function $f$ links one state in the trajectory to the next, given action $a_t$, via the Markovian rule $\vec{x}_{t+1} = f(\vec{x}_t, a_t)$. The second model function, $r$, gives an immediate real-valued reward $r_t = r(\vec{x}_t, a_t)$ on arriving at the next state $\vec{x}_{t+1}$.

Assume that each trajectory is guaranteed to reach a terminal state in some finite time (i.e. the problem is episodic). Note that in general, the number of time steps in a trajectory may be dependent on the actions taken. For example, a scenario like this could be an aircraft with limited fuel trying to land. For a particular trajectory label the final time step $t = F$, so that $\vec{x}_F$ is the terminal state of that trajectory. Assume each action $a_t$ is a real number that, for some problems, may be constrained to $-1 \leq a_t \leq 1$.

For any trajectory starting at state $\vec{x}_0$ and following actions $\{a_0, a_1, \ldots, a_{F-1}\}$ until reaching a terminal state under the given model, the total reward encountered is given by the function:

$$\begin{aligned} R(\vec{x}_0, a_0, a_1, \ldots, a_{F-1}) &= \sum_{t=0}^{F-1} r(\vec{x}_t, a_t) \\ &= r(\vec{x}_0, a_0) + R(f(\vec{x}_0, a_0), a_1, a_2, \ldots, a_{F-1}) \end{aligned} \tag{1}$$

Thus $R$ is a function of the arbitrary starting state $\vec{x}_0$ and the actions, and this allows us to obtain the partial derivative $\frac{\partial R}{\partial \vec{x}}$.

**Policy.** A policy is a function $\pi(\vec{x}, \vec{w})$, parametrised by weight vector $\vec{w}$, that generates actions as a function of state. Thus for a given trajectory generated by a given policy $\pi$, $a_t = \pi(\vec{x}_t, \vec{w})$. Since the policy is a pure function of $\vec{x}$ and $\vec{w}$, the policy is memoryless.

If a trajectory starts at state $\vec{x}_0$ and then follows a policy $\pi(\vec{x}, \vec{w})$ until reaching a terminal state, then the total reward is given by the function:

$$
\begin{aligned}
R^\pi(\vec{x}_0, \vec{w}) &= \sum_{t=0}^{F-1} r(\vec{x}_t, \pi(\vec{x}_t, \vec{w})) \\
&= r(\vec{x}_0, \pi(\vec{x}_0, \vec{w})) + R^\pi(f(\vec{x}_0, \pi(\vec{x}_0, \vec{w})), \vec{w})
\end{aligned}
$$

**Approximate Value Function.** We define $V(\vec{x}, \vec{w})$ to be the real-valued output of a smooth function approximator with weight vector $\vec{w}$ and input vector $\vec{x}$.[1] We refer to $V(\vec{x}, \vec{w})$ simply as the "value function" over state space $\vec{x}$, parametrised by weights $\vec{w}$.

**Q Value function.** The Q Value function (Watkins and Dayan, 1992) is defined as

$$
Q(\vec{x}, a, \vec{w}) = r(\vec{x}, a) + V(f(\vec{x}, a), \vec{w}) \tag{2}
$$

**Trajectory Shorthand Notation.** For a given trajectory through states $\{\vec{x}_0, \vec{x}_1, \ldots, \vec{x}_F\}$ with actions $\{a_0, a_1, \ldots, a_{F-1}\}$, and for any function defined on $S$ (e.g. including $V(\vec{x}, \vec{w})$, $G(\vec{x}, \vec{w})$, $R(\vec{x}, a_0, a_1, \ldots, a_{F-1})$, $R^\pi(\vec{x}, \vec{w})$, $r(\vec{x}, a)$, $V'(\vec{x}, \vec{w})$ and $G'(\vec{x}, \vec{w})$) we use a subscript of $t$ on the function to indicate that the function is being evaluated at $(\vec{x}_t, \vec{w})$. For example, $r_t = r(\vec{x}_t, a_t)$, $G_t = G(\vec{x}_t, \vec{w})$, $R^\pi{}_t = R^\pi(\vec{x}_t, \vec{w})$ and $R_t = R(\vec{x}_t, a_t, a_{t+1}, \ldots, a_{F-1})$. Note that this shorthand does not mean that these functions are functions of $t$, as that would break the Markovian condition.

Similarly, for any of these function's partial derivatives, we use brackets with a subscripted $t$ to indicate that the partial derivative is to be evaluated at time step $t$. For example, $\left(\frac{\partial G}{\partial w}\right)_t$ is shorthand for $\left.\frac{\partial G}{\partial w}\right|_{(\vec{x}_t, \vec{w})}$, i.e. the function $\frac{\partial G}{\partial w}$ evaluated at $(\vec{x}_t, \vec{w})$. Also, for example, $\left(\frac{\partial f}{\partial a}\right)_t = \left.\frac{\partial f}{\partial a}\right|_{(\vec{x}_t, a_t)}$; $\left(\frac{\partial R}{\partial \vec{x}}\right)_t = \left(\left.\frac{\partial R}{\partial \vec{x}}\right|_{(\vec{x}_t, a_t, a_{t+1}, \ldots, a_{F-1})}\right)$; and similarly for other partial derivatives including $\left(\frac{\partial r}{\partial \vec{x}}\right)_t$ and $\left(\frac{\partial R^\pi}{\partial \vec{w}}\right)_t$.

**Greedy Policy.** The greedy policy on $V$ generates $\pi(\vec{x}, \vec{w})$ such that

$$
\pi(\vec{x}, \vec{w}) = \arg\max_a (Q(\vec{x}, a, \vec{w})) \tag{3}
$$

subject to the constraints (if present) that $-1 \leq a_t \leq 1$. The greedy policy is a one-step look-ahead that decides which action to take, based only on the model and $V$. A *greedy trajectory* is one that has been generated by the greedy policy. Since for a greedy policy, the actions are dependent on the value function and state, and $V = V(\vec{x}, \vec{w})$, it follows that $\pi = \pi(\vec{x}, \vec{w})$. This means that any modification to the weight vector $\vec{w}$ will immediately change $V(\vec{x}, \vec{w})$ and move all greedy trajectories. Hence we say the value function and greedy policy are *tightly coupled*.

For the greedy policy, when the constraints $-1 \leq a_t \leq 1$ are present, we say an action $a_t$ is *saturated* if $|a_t| = 1$ and $\left(\frac{\partial Q}{\partial a}\right)_t \neq 0$. If either of these conditions is not met, or the constraints are not present, then $a_t$ is not saturated. We note two useful consequences of this:

---

1. This differs slightly from some definitions in the RL literature, which would refer to this use of the function $V$ as an approximated value function for the greedy policy on $V$. To side-step this circularity in definition, we have treated $V(\vec{x}, \vec{w})$ simply as a smooth function on which a greedy policy can be defined.

**Lemma 1** *If $a_t$ is not saturated, then $\left(\frac{\partial Q}{\partial a}\right)_t = 0$ and $\left(\frac{\partial^2 Q}{\partial a^2}\right)_t \leq 0$ (since it's a maximum).*

**Lemma 2** *If $a_t$ is saturated, then, whenever they exist, $\left(\frac{\partial \pi}{\partial \vec{x}}\right)_t = 0$ and $\left(\frac{\partial \pi}{\partial \vec{w}}\right)_t = 0$.*

Note that $\left(\frac{\partial \pi}{\partial \vec{x}}\right)_t$ and $\left(\frac{\partial \pi}{\partial \vec{w}}\right)_t$ may not exist, for example, if there are multiple joint maxima in $Q(\vec{x}, a, \vec{w})$ with respect to $a$.

**$\epsilon$-Greedy Policy.** In our later experiments we implement VL algorithms that require some exploration. Hence we make use of a slightly modified version of the greedy policy which we refer to as the $\epsilon$-greedy policy:[2]

$$\pi(\vec{x}, \vec{w}) = \arg\max_a(Q(\vec{x}, a, \vec{w})) + RND(\epsilon)$$

Here $RND(\epsilon)$ is defined to be a random number generator that returns a normally distributed random variable with mean 0 and standard deviation $\epsilon$.

**The Value-Gradient - definition.** The value-gradient function $G(\vec{x}, \vec{w})$ is the derivative of the value function $V(\vec{x}, \vec{w})$ with respect to state vector $\vec{x}$. Therefore $G(\vec{x}, \vec{w}) = \frac{\partial V(\vec{x}, \vec{w})}{\partial \vec{x}}$. Since $V(\vec{x}, \vec{w})$ is defined to be smooth, the value-gradient always exists.

**Targets for V.** For a trajectory found by a greedy policy $\pi(\vec{x}, \vec{w})$ on a value function $V(\vec{x}, \vec{w})$, we define the function $V'(\vec{x}, \vec{w})$ recursively as

$$V'(\vec{x}, \vec{w}) = r(\vec{x}, \pi(\vec{x}, \vec{w})) + \big(\lambda V'(f(\vec{x}, \pi(\vec{x}, \vec{w})), \vec{w}) + (1 - \lambda)V(f(\vec{x}, \pi(\vec{x}, \vec{w})), \vec{w})\big) \quad (4)$$

with $V'(\vec{x}_F, \vec{w}) = 0$ and where $0 \leq \lambda \leq 1$ is a fixed constant. To calculate $V'$ for a particular point $\vec{x}_0$ in state space, it is necessary to run and cache a whole trajectory starting from $\vec{x}_0$ under the greedy policy $\pi(\vec{x}, \vec{w})$, and then work backwards along it applying the above recursion; thus $V'(\vec{x}, \vec{w})$ is defined for all points in state space.

Using shorthand notation, the above equation simplifies to

$$V'_t = r_t + \big(\lambda V'_{t+1} + (1 - \lambda)V_{t+1}\big)$$

$\lambda$ is a "bootstrapping" parameter, giving full bootstrapping when $\lambda = 0$ and none when $\lambda = 1$, described by Sutton (1988). When $\lambda = 1$, $V'(\vec{x}, \vec{w})$ becomes identical to $R^\pi(\vec{x}, \vec{w})$. For any $\lambda$, $V'$ is identical to the "$\lambda$-return", or the "forward view of TD($\lambda$)", described by Sutton and Barto (1998). The use of $V'$ greatly simplifies the analysis of value functions and value-gradients.

We refer to the values $V'_t$ as the "targets" for $V_t$. The objective of any VL algorithm is to achieve $V_t = V'_t$ for all $t > 0$ along *all possible greedy* trajectories. By Eq. 4 and for any $\lambda$, this objective becomes equivalent to the deterministic and undiscounted case of the Bellman Equation of dynamic programming (Sutton and Barto, 1998):

$$V_t = V'_t \;\; \forall t > 0 \quad \Longleftrightarrow \quad V_t = r_t + V_{t+1} \;\; \forall t > 0 \quad (5)$$

Since the Bellman Equation needs satisfying at all points in state space, it is sometimes referred to as a *global* method, and this means VL algorithms always need to incorporate some form of exploration.

---

2. This differs from the definition of the $\epsilon$-greedy policy that Sutton and Barto (1998) use. Also in the definition we use, we have assumed the actions are unconstrained.

We point out that since $V'$ is dependent on the actions and on $V(\vec{x}, \vec{w})$, it is not a simple matter to attain the objective $V \equiv V'$, since changing $V$ infinitesimally will immediately move the greedy trajectories (since they are tightly coupled), and therefore change $V'$; these targets are moving ones. However, a learning algorithm that continually attempts to move the values $V_t$ infinitesimally and directly towards the values $V'_t$ is equivalent to TD($\lambda$) (Sutton, 1988), as shown in Section 1.2.1. This further justifies Eq. 4.

**Matrix-vector notation.** Throughout this paper, a convention is used that all defined vector quantities are columns, and any vector becomes transposed (becoming a row) if it appears in the numerator of a differential. Upper indices indicate the component of a vector. For example, $\vec{x}_t$ is a column; $\vec{w}$ is a column; $G_t$ is a column; $\left(\frac{\partial R^\pi}{\partial \vec{w}}\right)_t$ is a column; $\left(\frac{\partial f}{\partial a}\right)_t$ is a row; $\left(\frac{\partial f}{\partial \vec{x}}\right)_t$ is a matrix with element $(i,j)$ equal to $\left(\frac{\partial f(\vec{x},a)^j}{\partial \vec{x}^i}\right)_t$; $\left(\frac{\partial G}{\partial \vec{w}}\right)_t$ is a matrix with element $(i,j)$ equal to $\left(\frac{\partial G^j}{\partial \vec{w}^i}\right)_t$. An example of a product is $\left(\frac{\partial f}{\partial a}\right)_t G_{t+1} = \sum_i \left(\frac{\partial f^i}{\partial a}\right)_t G^i_{t+1}$.

**Target Value-Gradient, $(G'_t)$.** We define $G'(\vec{x}, \vec{w}) = \frac{\partial V'(\vec{x}, \vec{w})}{\partial \vec{x}}$. Expanding this, using Eq. 4, gives:

$$G'_t = \left(\left(\frac{\partial r}{\partial \vec{x}}\right)_t + \left(\frac{\partial \pi}{\partial \vec{x}}\right)_t \left(\frac{\partial r}{\partial a}\right)_t\right) + \left(\left(\frac{\partial f}{\partial \vec{x}}\right)_t + \left(\frac{\partial \pi}{\partial \vec{x}}\right)_t \left(\frac{\partial f}{\partial a}\right)_t\right)(\lambda G'_{t+1} + (1-\lambda)G_{t+1}) \tag{6}$$

with $G'_F = \vec{0}$. To obtain this total derivative we have used the fact that $a_t = \pi(\vec{x}_t, \vec{w})$, and that therefore changing $\vec{x}_t$ will immediately change all later actions and states.

This recursive formula takes a known target value-gradient at the end point of a trajectory ($G'_F = \vec{0}$), and works it backwards along the trajectory rotating and incrementing it as appropriate, to give the desired value-gradient at each time step. This is the central equation behind all VGL algorithms; the objective for any VGL algorithm is to attain $G_t = G'_t$ for all $t > 0$ along a greedy trajectory. As with the target values, it should be noted that this objective is not straightforward to achieve since the values $G'_t$ are moving targets and are highly dependent on $\vec{w}$.

The above objective $G \equiv G'$ is a *local* requirement that only needs satisfying along a greedy trajectory, and is usually sufficient to ensure the trajectory is locally optimal. This is in stark contrast to the Bellman Equation for VL (Eq. 5) which is a *global* requirement. Consequently VGL is potentially much more efficient and effective than VL. This difference is justified and explained further in Section 1.3.

All terms of Eq. 6 are obtainable from knowledge of the model functions and the policy. For obtaining the term $\frac{\partial \pi}{\partial \vec{x}}$ it is usually preferable to have the greedy policy written in analytical form, as done in Section 2.2 and the experiments of Section 4. Alternatively, using a derivation similar to that of Eq. 17, it can be shown that, when it exists,

$$\left(\frac{\partial \pi}{\partial \vec{x}}\right)_t = \begin{cases} -\left(\frac{\partial^2 Q}{\partial \vec{x} \partial a}\right)_t \left(\frac{\partial^2 Q}{\partial a^2}\right)_t^{-1} & \text{if } a_t \text{ unsaturated and } \left(\frac{\partial^2 Q}{\partial a^2}\right)_t^{-1} \text{ exists} \\ 0 & \text{if } a_t \text{ saturated} \end{cases}$$

If $\lambda > 0$ and if $\frac{\partial \pi}{\partial \vec{x}}$ does not exist at some time step, $t_0$, of the trajectory, then $G'_t$ is not defined for all $t \leq t_0$. In some common situations, such as the continuous-time formulations (Section 2.2), $\frac{\partial \pi}{\partial \vec{x}}$ is always defined so this is not a problem.

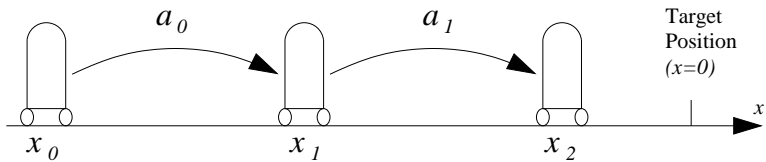

Figure 1: Illustration of 2-step Toy Problem

In the special case where $\lambda = 0$, the dependency of Eq. 6 on $\frac{\partial \pi}{\partial \vec{x}}$ disappears and so $G'_t$ always exists, and the definition reduces to $G'_t = \left(\frac{\partial r}{\partial \vec{x}}\right)_t + \left(\frac{\partial f}{\partial \vec{x}}\right)_t G_{t+1}$. In this case $G'_t$ is equivalent to the target value-gradient that Werbos uses in the algorithms DHP and GDHP (Werbos, 1998, Eq. 18).

When $\lambda = 1$, $G'_t$ becomes identical to $\left(\frac{\partial R^\pi}{\partial \vec{x}}\right)_t$. $\left(\frac{\partial R^\pi}{\partial \vec{x}}\right)_t$ appears explicitly in the equations of differential dynamic programming (Jacobson and Mayne, 1970), and implicitly in back-propagation through time (Eq. 15).

### 1.1.1 Example - Toy Problem

Many experiments in this paper make use of the *n-step Toy Problem with parameter $k$*. This is a problem in which an agent can move along a straight line and must move towards the origin efficiently in a given number of time steps, illustrated in Fig. 1, and defined generically now:

State space is one-dimensional and continuous. The actions are unbounded. In this episodic problem, we define the model functions differently at each time step, and each trajectory is defined to terminate at time step $t = n + 1$. Strictly speaking, to satisfy the Markovian requirement and achieve these time step dependencies, we should add one extra dimension to state space to hold $t$ and adjust the model functions accordingly. However this complication was omitted in the interests of keeping the notation simple. Under this simplification, the model functions are:

$$f(x_t, a_t) = \begin{cases} x_t + a_t & \text{if } 0 \le t < n \\ x_t & \text{if } t = n \end{cases} \tag{7a}$$

$$r(x_t, a_t) = \begin{cases} -k{a_t}^2 & \text{if } 0 \le t < n \\ -{x_t}^2 & \text{if } t = n \end{cases} \tag{7b}$$

where $k$ is a real-valued non-negative constant to allow more varieties of problem types to be specified. The $(n+1)^{th}$ time step is present just to deliver a final reward based only on state. The model functions in the time step $t = n$ are independent of the action $a_n$, and so each trajectory is completely parametrised by just $(x_0, a_0, a_1, \ldots, a_{n-1})$. This completes the definition of the Toy Problem.

Next we describe the optimal trajectories and optimal policy for the $n$-step Toy Problem. Since the total reward is

$$R(x_0, a_0, a_1, \ldots, a_{n-1}) = -k({a_0}^2 + {a_1}^2 + \ldots + {a_{n-1}}^2) - (x_n)^2$$
$$= -k({a_0}^2 + {a_1}^2 + \ldots + {a_{n-1}}^2) - (x_0 + a_0 + a_1 + \ldots + a_{n-1})^2$$

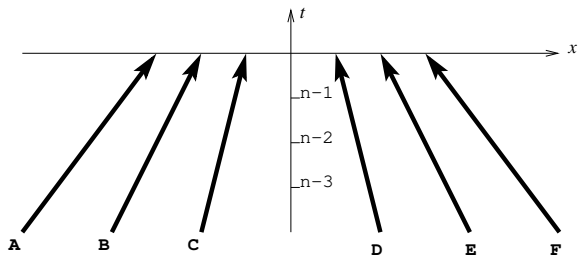

Figure 2: The lines A to F are optimal trajectories for the Toy Problem. The trajectories are shown as continuous lines for illustration here, although time is defined to be discrete in this problem. The arrowheads of the trajectories show the position of the terminal states.

then the actions that maximise this are all equal, and are given by,

$$a_t = -\frac{x_0}{n+k} \text{ for all } 0 \le t < n.$$

Since the optimal actions are all equal and directed oppositely to the initial state $x_0$, optimal trajectories form straight lines towards the centre as shown in Figure 2. Each optimal trajectory terminates at $x_n = \frac{kx_0}{n+k}$. The optimal policy, usually denoted $\pi^*(\vec{x}_t)$, is

$$\pi^*(x_t) = -\frac{x_t}{n-t+k} \text{ for all } 0 \le t < n \tag{8}$$

The optimal value function, usually denoted $V^*(\vec{x}_t)$, can be found for this problem by evaluating the total reward encountered on following the optimal policy until termination:

$$V^*(x_t) = \begin{cases} -\frac{k(x_t)^2}{n-t+k} & \text{if } t \le n \\ 0 & \text{if } t = n+1 \end{cases} \tag{9}$$

For a simple example of a value function and greedy policy, the reader should see Appendix B (equations 33 and 34).

## 1.2 Value-Learning Methods

Introducing the targets $V'$ simplifies the formulation of some learning algorithms. The objective of all VL algorithms is to learn $V_t = V'_t$ for all $t \ge 1$. We do not need to consider the time step $t = 0$ since the greedy policy is independent of the value function at $t = 0$. A quick review of two common VL methods follows.

All learning algorithms in this paper are "off-line", that is any weight updates are applied only after considering a whole trajectory. In all learning algorithms we take $\alpha$ to be a small positive constant.

### 1.2.1 TD($\lambda$) Learning

The TD($\lambda$) algorithm (Sutton, 1988) attempts to achieve $V_t = V'_t$ for all $t \geq 1$ by the following weight update:

$$\Delta \vec{w} = \alpha \sum_{t \geq 1} \left( \frac{\partial V}{\partial \vec{w}} \right)_t (V'_t - V_t) \tag{10}$$

The equivalence of this formulation to that used by Sutton (1988) is proven in Appendix C. This equivalence validates the $V'$ notation. Although not originally defined for control problems by Sutton (1988), it is shown in Appendix D that the TD($\lambda$) weight update can be applied directly to control problems with a known model using the $\epsilon$-greedy policy, and is then equivalent to Sarsa($\lambda$) (Rummery and Niranjan, 1994).

Unfortunately there are no convergence guarantees for this equation when used with a general function approximator for the value function, and divergence examples abound.

As shown by Sutton and Barto (1998) TD learning becomes Monte-Carlo learning when the bootstrapping parameter, $\lambda = 1$.

### 1.2.2 Residual Gradients

With the aim of improving the convergence guarantees of the previous method, the approach of Baird (1995) is to minimize the error $E = \frac{1}{2} \sum_{t \geq 1} (V'_t - V_t)^2$ by gradient descent on the weights:

$$\Delta \vec{w} = \alpha \sum_{t \geq 1} \left( \left( \frac{\partial V}{\partial \vec{w}} \right)_t - \left( \frac{\partial V'}{\partial \vec{w}} \right)_t \right) (V'_t - V_t)$$

The extra terms introduced by this method are referred to throughout this paper as the "residual gradient terms". We extend the residual gradient method to value-gradients in Section 2.3, and extend it to work with a greedy policy that is tightly coupled with the value function.

## 1.3 Motivation for Value-Gradients

The above VL algorithms need to use some form of exploration. The required exploration could be implemented by randomly varying the start point in state space at each iteration, a technique known as "exploring starts". Alternatively a stochastic model or policy could be used to force exploration within a single trajectory. Exploration introduces inefficiencies which are discussed in Section 1.5.

Figure 3 demonstrates why VL without exploration can lead to suboptimal trajectories, whereas VGL will not. Understanding this is a very central point to this paper as it is the central motivation for using value-gradients. If a fuller explanation of either of these cases is required, then Appendix B gives details, and goes further in showing that the trajectory with learned value-gradients will also be optimal.

The conclusion of this example, and Appendix B, is that without exploration, VL will possibly, and in fact be likely to, converge to a suboptimal trajectory. Exploration is necessary since it enables the learning algorithm to become aware of the presence of any superior neighbouring trajectories (which is exactly the information that a value-gradient contains). Without this awareness learning can terminate on a suboptimal trajectory, as this counterexample shows.

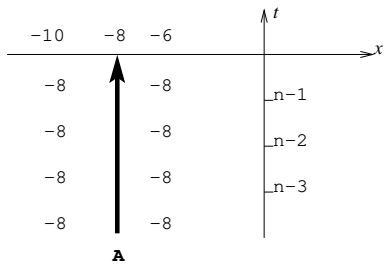

Figure 3: Illustration of the problem of VL without exploration. In this diagram the floating negative numbers in the bottom left quadrant represent value function spot values (for a value function which is constant throughout state space). Hence, in the Toy Problem the greedy policy will choose zero actions, and the greedy trajectory (from A) is the straight line shown. The negative numbers above the $x$ axis give the final reward, $r_n$. Since the intermediate rewards $r_t$ are zero for all $t < n$, we have $V'_t = -8$ for all $t \leq n$. So $V_t = V'_t = -8$ for all $t \leq n$, and so VL is complete; yet the trajectory is not optimal (c.f. Fig. 2). This situation cannot happen with VGL, since if the value-gradients were learned then there would be a value-gradient perpendicular to the trajectory, and the greedy policy would not have chosen the trajectory shown.

Note that this requirement for exploration is separate from the issue of exploration that is sometimes also required to learn an unknown model.

Similar counterexamples can be constructed for other problem domains. For example, Figure 3 can be applied to any problem where all the reward occurs at the end of an episode.

Furthermore, it is proven in Appendix A that in a general problem, learning the value-gradients along a greedy trajectory is a sufficient condition for the trajectory to be locally extremal, and also often ensures the trajectory is locally optimal (these terms are defined in the same appendix). This contrasts with VL in that there has been no need for exploration. *Local exploration comes for free with value-gradients, since knowledge of the value-gradients automatically provides knowledge of the neighbouring trajectories.*

## 1.4 Relationship of Value-Learning to Value-Gradient Learning

Learning the value-gradients along a trajectory learns the relative values of the value function along it and all of its immediately neighbouring trajectories. We refer to all these trajectories collectively as a *tube*. Any VL algorithm that exhibits sufficient exploration to learn the target values fully throughout the entire tube would also achieve this goal, and therefore also achieve locally optimal trajectories.

This shows the consistency of the two methods, and the equivalence of their objectives. *We believe VGL techniques represent an idealised form of VL techniques, and that VL is a stochastic approximation to VGL.* Both have similar objectives, i.e. to achieve an optimal trajectory by learning the relative target values throughout a tube; but whereas VL techniques rely on the scattergun approach of stochastic exploration to achieve this, VGL techniques go about it more methodically and directly.

Figure 4 illustrates the contrasting approaches of the two methods.

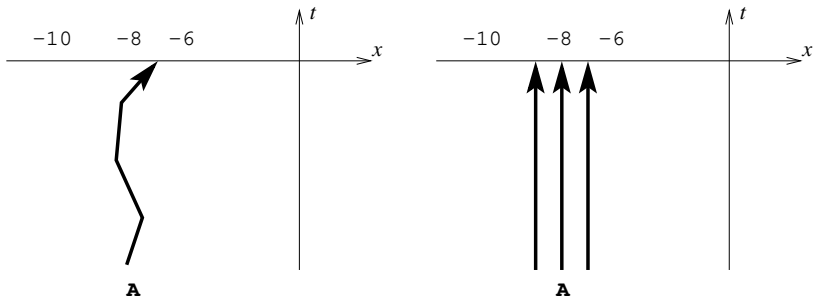

Figure 4: Diagrams contrasting VL by stochastic control (left figure) against deterministic learning by VGL (right figure). In the VL case (left figure), stochastic control makes the trajectory zigzag, and the target value passed backwards along the trajectory (which in this case is approximately $-7$, whereas the deterministic value should be $-8$) will be passed back to the points in state space encountered along the trajectory. In the VGL case (right figure), the value-gradient at the end point (which will be a small vector pointing in the positive $x$ direction) is passed backwards, without any stochastic distortion, along the central trajectory and therefore influences the value function along all three trajectories simultaneously.

Once the effects of exploration are averaged out in VL, the underlying weight update is often very similar to a VGL weight update; but possibly with some extra, and usually unwanted, terms. See Section 4.1 for an example analysis in a specific problem. Hence, we would expect a large proportion of results obtained for one method to apply to the other. For example, using a value-gradient analysis, in Section 4.3 we derive an example that causes VGL to diverge, and then empirically find this example causes divergence in VL too. Also, we would expect the analyses on residual gradients and actor-critic architectures to place limitations on what can be achieved with these methods when used with VL (sections 2.3 and 3).

If the value-gradient is learned throughout the whole of state space, i.e. if $\frac{\partial V}{\partial \vec{x}} = \frac{\partial V'}{\partial \vec{x}}$ for all $\vec{x}$, then this forms a differential equation of which the Bellman Equation ($V = V'$ for all $\vec{x}$; see Eq. 5) is a particular solution. This is illustrated in Equation 11.

$$G(\vec{x}, \vec{w}) = G'(\vec{x}, \vec{w}) \ \ \forall \vec{x} \quad \Longleftrightarrow \quad V(\vec{x}, \vec{w}) = V'(\vec{x}, \vec{w}) + c \ \ \forall \vec{x} \tag{11}$$

Of course the arbitrary constant, $c$, is not important as it does not affect the greedy policy. Hence we propose that *the values themselves are not important at all; it is only the value-gradients that are important*. This extreme view is consistent with the fact that it is value-gradients, not values, that appear in the optimality proof (Appendix A), in the relationship of value function learning to PGL (Section 2.1) and in the equation for $\frac{\partial \pi}{\partial \vec{w}}$ (Eq. 17).

The directness of the VGL method means it is more open to theoretical analysis than the VL method. The greedy policy is dependent on the value-gradient but not on the values (see Eq. 17 or Eq. 34), so it seems essential to consider value-gradients if we are to understand how an update to $\vec{w}$ will affect a greedy trajectory. If we define a "whole system" to mean the conjunction of a tightly coupled value function with a greedy policy, then *it is necessary to consider value-gradients if we are to understand the overall convergence properties of any "whole system" weight-update.*

### 1.5 Efficiencies of learning Value-Gradients

VGL introduces some significant efficiency improvements into value function learning.

- The removal of the need for exploration should be a major efficiency gain. As learning algorithms are already iterative, the need to explore neighbouring trajectories causes a nested layer of iteration. Also, the issue of exploration severely restricts standard algorithms, particularly those that work by learning the $Q(\vec{x}, a, \vec{w})$ function (e.g. Sarsa($\lambda$), Q($\lambda$)-learning (Watkins, 1989)). For example, whenever an action $a_t$ in a trajectory is exploratory, i.e. non-greedy, the target values backed up to all previous time steps (i.e. the values $V'_k$ for all $k < t$) will be changed. This effectively sends the wrong learning targets back to the early time steps. This difficulty is dealt with in Sarsa($\lambda$) by making $\epsilon$ slowly tend to zero in the $\epsilon$-greedy policy. This difficulty is dealt with in Q($\lambda$)-learning by forcing $\lambda = 0$ or truncating learning beyond the exploratory time step. Both of these solutions have performance implications (we show in Section 4.1 that as $\epsilon \to 0$, learning grinds to a halt).

- Learning the value-gradients along a trajectory is similar to learning the value function's relative values along an entire group of immediately neighbouring trajectories (see Fig. 4). Thus the value-gradients encapsulate more relevant information than the values, and therefore learning by value-gradients should be faster (provided the function approximator for $V$ can be made to learn gradients efficiently, which is not a trivial problem).

- As described in Section 2.2, some VGL algorithms are doing true gradient ascent on $R^\pi$, so are viable to speed up through any of the fast neural-network optimisation algorithms available.

These informal arguments are backed up by the experiments in Section 4.

## 2. Learning Algorithms for Value-Gradients

The objective of any VGL algorithm is to ensure $G_t = G'_t$ for all $t > 0$ along a greedy trajectory. As proven in Appendix A, this will be sufficient to ensure a locally extremal trajectory (and often a locally optimal trajectory). This section looks at some learning algorithms that try to achieve this objective. In Section 2.2 we derive a VGL algorithm that is guaranteed to converge to a locally optimal trajectory.

Define the sum-of-squares error function for value-gradients as

$$E(\vec{x}_0, \vec{w}) = \frac{1}{2} \sum_{t \geq 1} (G_t - G'_t)^T \Omega_t (G_t - G'_t) \tag{12}$$

for a given greedy trajectory. $\Omega_t$ is any arbitrarily chosen positive semi-definite matrix (as introduced by Werbos, 1998), included for generality, and often just taken to be the identity matrix for all $t$. A use for $\Omega_t$ could be to allow us to compensate explicitly for any rescaling of the dimensions of state-space. If bootstrapping is used then $\Omega_t$ should be chosen to be strictly positive definite.

One approach for VGL is to perform gradient descent on the above error function (giving the VGL counterpart to residual-gradients):

$$\Delta \vec{w} = \alpha \frac{\partial E}{\partial \vec{w}} \tag{13}$$

This equation is analysed in Section 2.3. However a simpler weight update is to omit the "residual gradient" terms (giving the VGL counterpart to TD($\lambda$)):

$$\Delta \vec{w} = \alpha \sum_{t \geq 1} \left( \frac{\partial G}{\partial \vec{w}} \right)_t \Omega_t (G'_t - G_t) \tag{14}$$

In the next section we prove that this equation, with $\lambda = 1$, is equivalent to PGL, and show it leads to a successful algorithm with convergence guarantees.

Any VGL algorithm is going to involve using the matrices $\left( \frac{\partial G}{\partial \vec{w}} \right)_t$ and/or $\frac{\partial G}{\partial \vec{x}}$ which, for neural-networks, involves second order back-propagation. This is described by (White and Sofge, 1992, ch. 10) and (Coulom, 2002, Appendix A). In fact, these matrices are only required when multiplied by a column vector, which can be implemented efficiently by extending the techniques of Pearlmutter (1994) to this situation.

## 2.1 Relationship to Policy-Gradient Learning

We now prove that the VGL update of Eq. 14, with $\lambda = 1$ and a carefully chosen $\Omega_t$ matrix, is equivalent to PGL on a greedy policy. It is this equivalence that provides convergence guarantees for Eq. 14. To make this demonstration clearest, it is easiest to start by considering a PGL weight update; although it should be pointed out that the discovery of this equivalence occurred the opposite way around, since forms of Eq. 14 date back prior to Werbos (1998).

PGL, sometimes also known as "direct" reinforcement learning, is defined to be gradient ascent on $R^\pi(\vec{x}_0, \vec{w})$ with respect to the weight vector $\vec{w}$ of the policy:

$$\Delta \vec{w} = \alpha \left( \frac{\partial R^\pi}{\partial \vec{w}} \right)_0$$

Back-propagation through time (BPTT) is merely an efficient implementation of this formula, designed for architectures where the policy $\pi(\vec{x}, \vec{w})$ is provided by a neural-network (see Werbos, 1990). PGL methods will naturally find stationary points that are constrained locally optimal trajectories (see Appendix A for optimality definitions).

For PGL, we therefore have:

$$
\begin{aligned}
\left(\frac{\partial R^\pi}{\partial \vec{w}}\right)_t &= \frac{\partial(r(\vec{x}_t, \pi(\vec{x}_t, \vec{w})) + R^\pi(f(\vec{x}_t, \pi(\vec{x}_t, \vec{w})), \vec{w}))}{\partial \vec{w}} \\
&= \left(\frac{\partial \pi}{\partial \vec{w}}\right)_t \left(\left(\frac{\partial r}{\partial a}\right)_t + \left(\frac{\partial f}{\partial a}\right)_t \left(\frac{\partial R^\pi}{\partial \vec{x}}\right)_{t+1}\right) + \left(\frac{\partial R^\pi}{\partial \vec{w}}\right)_{t+1} \\
\Rightarrow \Delta \vec{w} = \alpha \left(\frac{\partial R^\pi}{\partial \vec{w}}\right)_0 &= \alpha \sum_{t \geq 0} \left(\frac{\partial \pi}{\partial \vec{w}}\right)_t \left(\left(\frac{\partial r}{\partial a}\right)_t + \left(\frac{\partial f}{\partial a}\right)_t \left(\frac{\partial R^\pi}{\partial \vec{x}}\right)_{t+1}\right) \quad (15)
\end{aligned}
$$

This equation is identical to the weight update performed by BPTT. It is defined for a general policy. We now switch to specifically consider PGL applied to a greedy policy. Initially we only consider the case where $\left(\frac{\partial R^\pi}{\partial \vec{w}}\right)_0$ exists for a greedy trajectory, and hence $\left(\frac{\partial \pi}{\partial \vec{w}}\right)_t$ exists for all $t$. Now in the summation of Eq. 15, we only need to consider the time steps where $a_t$ is not saturated, since for $a_t$ saturated, $\left(\frac{\partial \pi}{\partial \vec{w}}\right)_t = 0$ (by Lemma 2).

The summation involves terms $\left(\frac{\partial \pi}{\partial \vec{w}}\right)_t$ and $\left(\frac{\partial r}{\partial a}\right)_t$ which can be reinterpreted under the greedy policy:

**Lemma 3** *The greedy policy implies, for an unsaturated action,*

$$
\begin{aligned}
\left(\frac{\partial Q}{\partial a}\right)_t &= \left(\frac{\partial r}{\partial a}\right)_t + \left(\frac{\partial f}{\partial a}\right)_t \left(\frac{\partial V}{\partial \vec{x}}\right)_{t+1} = 0 \\
\Rightarrow \left(\frac{\partial r}{\partial a}\right)_t &= -\left(\frac{\partial f}{\partial a}\right)_t G_{t+1} \quad (16)
\end{aligned}
$$

**Lemma 4** *When $\left(\frac{\partial \pi}{\partial \vec{w}}\right)_t$ exists for an unsaturated action $a_t$, the greedy policy implies $\left(\frac{\partial Q}{\partial a}\right)_t \equiv 0$ therefore,*

$$
\begin{aligned}
0 &= \frac{\partial}{\partial \vec{w}} \left(\frac{\partial Q(\vec{x}_t, \pi(\vec{x}_t, \vec{w}), \vec{w})}{\partial a_t}\right) = \left(\frac{\partial}{\partial \vec{w}} + \left(\frac{\partial \pi}{\partial \vec{w}}\right)_t \frac{\partial}{\partial a_t}\right) \left(\frac{\partial Q(\vec{x}_t, a_t, \vec{w})}{\partial a_t}\right) \\
&= \frac{\partial}{\partial \vec{w}} \left(\left(\frac{\partial r}{\partial a}\right)_t + \left(\frac{\partial f}{\partial a}\right)_t G(\vec{x}_{t+1}, \vec{w})\right) + \left(\frac{\partial \pi}{\partial \vec{w}}\right)_t \left(\frac{\partial^2 Q}{\partial a^2}\right)_t \\
&= \left(\frac{\partial G}{\partial \vec{w}}\right)_{t+1} \left(\frac{\partial f}{\partial a}\right)_t^T + \left(\frac{\partial \pi}{\partial \vec{w}}\right)_t \left(\frac{\partial^2 Q}{\partial a^2}\right)_t \\
\Rightarrow \left(\frac{\partial \pi}{\partial \vec{w}}\right)_t &= -\left(\frac{\partial G}{\partial \vec{w}}\right)_{t+1} \left(\frac{\partial f}{\partial a}\right)_t^T \left(\frac{\partial^2 Q}{\partial a^2}\right)_t^{-1}, \text{ assuming } \left(\frac{\partial^2 Q}{\partial a^2}\right)_t \neq 0 \quad (17)
\end{aligned}
$$

It now becomes possible to analyse the PGL weight update with a greedy policy. Substituting the results of the above lemmas (Eq. 16 and Eq. 17), and $\left(\frac{\partial R^\pi}{\partial \vec{x}}\right)_t = G'_t$ with

$\lambda = 1$ (see Eq. 6), into Eq. 15 gives:

$$
\begin{aligned}
\Delta \vec{w} &= \alpha \left( \frac{\partial R^\pi}{\partial \vec{w}} \right)_0 \\
&= \alpha \sum_{t \geq 0} \left( -\left( \frac{\partial G}{\partial \vec{w}} \right)_{t+1} \left( \frac{\partial f}{\partial a} \right)_t^T \left( \frac{\partial^2 Q}{\partial a^2} \right)_t^{-1} \left( \frac{\partial f}{\partial a} \right)_t (-G_{t+1} + G'_{t+1}) \right) \\
&= \alpha \sum_{t \geq 0} \left( \frac{\partial G}{\partial \vec{w}} \right)_{t+1} \Omega_t (G'_{t+1} - G_{t+1})
\end{aligned}
\tag{18}
$$

where

$$
\Omega_t = -\left( \frac{\partial f}{\partial a} \right)_t^T \left( \frac{\partial^2 Q}{\partial a^2} \right)_t^{-1} \left( \frac{\partial f}{\partial a} \right)_t,
\tag{19}
$$

and is positive semi-definite, by the greedy policy (Lemma 1).

Equation 18 is identical to a VGL weight update equation (Eq. 14), with a carefully chosen matrix for $\Omega_t$, and $\lambda = 1$, provided $\left( \frac{\partial \pi}{\partial \vec{w}} \right)_t$ and $\left( \frac{\partial^2 Q}{\partial a^2} \right)_t^{-1}$ exist for all $t$. If $\left( \frac{\partial \pi}{\partial \vec{w}} \right)_t$ does not exist, then $\frac{\partial R^\pi}{\partial \vec{w}}$ is not defined either. Resolutions to these existence conditions are proposed at the end of this section.

This completes the demonstration of the equivalence of a VGL algorithm (with the conditions stated above) to PGL (with greedy policy; when $\frac{\partial R^\pi}{\partial \vec{w}}$ exists) . Unfortunately we couldn't find a similar analysis for $\lambda < 1$, and divergence examples in this case are given in Section 4.3.

This result for $\lambda = 1$ was quite a surprise. It justifies the omission of the "residual gradient terms" when forming the weight update equation (Eq. 14). Omitting these residual gradient terms is not, as it may have seemed, a puzzling modification to $\frac{\partial E}{\partial \vec{w}}$; it is really $\frac{\partial R^\pi}{\partial \vec{w}}$, (with $\lambda = 1$, and the given $\Omega_t$). This means using the $\Omega_t$ terms ensures an optimal value of $R^\pi$ is obtained, as shown in the experiment in Section 4.4. Also, it shows that VGL algorithms (and hence value function learning algorithms) are not that different from PGL algorithms after all. It was not known that a PGL weight update, when applied to a greedy policy on a value function, would be doing the same thing as a value function weight update; even if both had $\lambda = 1$. Of course they are usually *not* the same, unless this particular choice of $\Omega_t$ is chosen.

This also provides a tentative justification for the TD($\lambda$) weight update equation (Eq. 10). From the point of view of the author, this previously had no theoretical justification. It was seemingly chosen because it looks a bit like gradient descent on an error function, and the Bellman Equation happens to be a fixed point of it. This has been a hugely puzzling issue. There are no convergence guarantees for it and numerous divergence examples (in Section 4.3, we show it can even diverge with $\lambda = 1$). Our explanation for it is that it is a stochastic approximation to Eq. 14, which itself is an approximation to PGL when $\lambda = 1$.

Also it is our understanding that this is a particularly good form of VGL weight update to make, since it has good convergence guarantees. If an alternative is chosen, e.g. by replacing $\Omega_t$ by the identity matrix, then it might be possible to get much more aggressive learning.[3] TD($\lambda$), being a stachostic approximation to Eq. 14, is fixed to implicitly use an identity matrix for $\Omega_t$. But this creates the unwanted problem of non-monotonic progress, in the same way that any aggressive modification to gradient ascent may do. It is also possible to get divergence in this case (see Section 4.3). It is our opinion that it is better to use a more theoretically justifiable acceleration method such as conjugate-gradients or RPROP.

This equivalence reduces the possible advantages of the value function architecture, in the case of $\lambda = 1$, down to being solely a sophisticated implementation of a policy function. This sophisticated policy architecture may just be easier to train than other policies; just as some neural-network architectures are easier to train than others. It is not the actual learning algorithm that is delivering any benefits.

We note that it also creates a difficulty in distinguishing between PGL and VGL. With $\lambda = 1$ and $\Omega_t$ as defined in Eq. 19, the equation can no longer be claimed as a new learning algorithm, since it is the same as BPTT with a greedy policy. Therefore the experimental results will be exactly the same as for BPTT. However, we will describe the above weight update as a VGL update; it is of the same form as Eq. 14. We also point out that forms of Eq. 14 came first (see Werbos, 1998), before the connection to PGL was realised, and that it itself is an idealised form of the TD($\lambda$) weight update.

This equivalence proof is almost a convergence proof for Eq. 18 with $\lambda = 1$, since for the majority of learning iterations there is smooth gradient ascent on $R^\pi$. The problem is that sometimes the terms $\left(\frac{\partial \pi}{\partial \vec{w}}\right)_t$ are not defined and then learning progress jumps discontinuously. One solution to this problem could be to choose a function approximator for $V$ such that the function $Q(\vec{x}, a, \vec{w})$ is everywhere strictly concave with respect to $a$, as is done in the Toy Problem experiments of Section 4. A more general solution is given in the next section. Both of these solutions also satisfy the requirement that $\left(\frac{\partial^2 Q}{\partial a^2}\right)_t^{-1}$ exists for all $t$.

## 2.2 Continuous-Time Formulation

In many problems time can be treated as a continuous variable, i.e. $t \in [0, F]$, as considered by Doya (2000) and Baird (1994). With continuous-time formulations, some extra difficulties can arise for VL as described and solved by Baird (1994), but these do not apply to VGL, for reasons described further below. We describe a continuous-time formulation here, since, in some circumstances the greedy policy $\pi(\vec{x}, \vec{w})$ becomes a smooth function of $\vec{x}$ and $\vec{w}$. This removes the problem of undefined $\left(\frac{\partial \pi}{\partial \vec{w}}\right)_t$ terms that was described in the previous section, and leads to a VGL algorithm for control problems with a function approximator that is guaranteed to converge.

We use bars over the previously defined functions to denote their continuous-time counterparts, so that $\bar{f}(\vec{x}, a)$ and $\bar{r}(\vec{x}, a)$ denote the continuous-time model functions. The trajectory is generated from a given start point $\vec{x}_0$ by the differential equation (DE),

$$\frac{d\vec{x}_t}{dt} = \bar{f}(\vec{x}_t, \pi(\vec{x}_t, \vec{w})) \tag{20}$$

---

3. In fact, in the continuous-time formulation of Eq. 23, $\left(\frac{\partial^2 Q}{\partial a^2}\right)_t^{-1} = -g'\left(\left(\frac{\partial \bar{r}^L}{\partial a}\right)_t + \left(\frac{\partial \bar{f}}{\partial a}\right)_t G_t\right)$, and so setting $\Omega_t$ to the identity matrix is analogous to giving the derivative of a sigmoid function an artificial boost (see Eq. 19). This is like the trick proposed by Fahlman (1988) that is sometimes used to speed up supervised learning in artificial neural networks, but at the expense of robustness.

The total reward for this trajectory is

$$R^\pi(\vec{x}_0, \vec{w}) = \int_0^F \bar{r}(\vec{x}_t, a_t)dt$$

The continuous-time $Q$ function is

$$\bar{Q}(\vec{x}, a, \vec{w}) = \bar{r}(\vec{x}, a) + \bar{f}^T(\vec{x}, a)G(\vec{x}, \vec{w}) + V(\vec{x}, \vec{w}) \tag{21}$$

This is closely related to the "advantage function" (Baird, 1994), that is $\bar{A}(\vec{x}, a, \vec{w}) = \bar{Q}(\vec{x}, a, \vec{w}) - V(\vec{x}, \vec{w})$. The major difference of the VGL approach over advantage functions is that "advantage learning" only learns the component of $G$ that is parallel to the trajectory, and so it is similar to all VL algorithms in that constant exploration of the neighbouring trajectories must take place. Also, as pointed out by Doya (2000), the problem of indistinguishable $Q$ values that advantage-learning is designed to address is not relevant when using the following policy:

We use the same policy as proposed by Doya (2000). The greedy policy does not need to look ahead in the continuous-time formulation, and instead relies only on the value-gradient at the current time. We assume the model functions are linear in $a$ (which is common in Newtonian models, since acceleration is proportional to the force), and then we introduce an extra "action-cost" non-linear term, $\bar{r}^C(\vec{x}, a)$, into the model's reward function, giving

$$\bar{r}(\vec{x}, a) = \bar{r}^L(\vec{x}, a) + \bar{r}^C(\vec{x}, a) \tag{22}$$

where $\bar{r}^L(\vec{x}, a)$ is the original linear reward function. The action-cost term has the effect of ensuring the action chosen by the greedy policy is bound to $[-1, 1]$, and also that the actions $a_t$ are smooth functions of $G_t$. A suitable choice is $\bar{r}^C(\vec{x}, a) = -\int_0^a g^{-1}(x)dx$ where $g^{-1}$ is the inverse of $g(x) = \tanh(x/c)$, and $c$ is a positive constant. This idea is explained more fully by Doya (2000).

Using this choice of $\bar{r}^C(\vec{x}, a)$, and substituting Eq. 21 and Eq. 22 into the greedy policy gives:

$$
\begin{aligned}
a_t = \pi(\vec{x}_t, \vec{w}) &= g\left(\left(\frac{\partial \bar{r}^L}{\partial a}\right)_t + \left(\frac{\partial \bar{f}}{\partial a}\right)_t G_t\right) \\
\Rightarrow \left(\frac{\partial \pi}{\partial \vec{w}}\right)_t &= g'\left(\left(\frac{\partial \bar{r}^L}{\partial a}\right)_t + \left(\frac{\partial \bar{f}}{\partial a}\right)_t G_t\right)\left(\frac{\partial G}{\partial \vec{w}}\right)_t \left(\frac{\partial \bar{f}}{\partial a}\right)_t^T \\
&= \frac{1 - a_t^2}{c}\left(\frac{\partial G}{\partial \vec{w}}\right)_t \left(\frac{\partial \bar{f}}{\partial a}\right)_t^T \\
\text{and } \left(\frac{\partial \pi}{\partial \vec{x}}\right)_t &= \frac{1 - a_t^2}{c}\left(\frac{\partial \bar{f}}{\partial \vec{x}}\right)_t \left(\frac{\partial G}{\partial \vec{x}}\right)_t \left(\frac{\partial \bar{f}}{\partial a}\right)_t^T
\end{aligned}
\tag{23}
$$

For this policy, as $c \to 0$, the policy tends to "bang-bang" control. For $c > 0$, this policy function meets the objective of producing bound actions that are smooth functions of $G_t$, and therefore, since the function approximator is assumed smooth, are also smooth functions of $\vec{w}$. This solves the problem of discontinuities described in the previous section.

The solution works for any $c > 0$, so can get arbitrarily close to the ideal of bang-bang control.

Using this policy, the trajectory can be calculated via Eq. 20 using a suitable DE solver. For small $c$, the actions can rapidly alternate and so the DE may be "stiff" and need an appropriate solver (see Press et al., 1992, ch.16.6).

For the learning equations in continuous-time, we use $\bar{\lambda}$ as the 'bootstrapping' parameter. This is related to the discrete time $\lambda$ by $e^{-\bar{\lambda}\Delta t} = \lambda$ where $\Delta t$ is the discrete-time time-step. This means $\bar{\lambda} = 0$ gives no bootstrapping, and that bootstrapping increases as $\bar{\lambda} \to \infty$, i.e. this is the opposite way around to the discrete-time parameter $\lambda$.

The equations in the rest of this section were derived in a similar manner to the discrete-time case, and by letting the discrete time-step $\Delta t \to 0$.

The continuous-time target-value has several different equivalent formulations:

$$
\begin{aligned}
V'_{t_0} &= \int_{t_0}^{F} e^{-\bar{\lambda}(t-t_0)} \left( \bar{r}_t + \frac{\partial V_t}{\partial t} \right) dt + V_{t_0} = \int_{t_0}^{F} e^{-\bar{\lambda}(t-t_0)} \left( \bar{r}_t + \bar{\lambda} V_t \right) dt \\
\Rightarrow \frac{\partial V'_t}{\partial t} &= -\bar{r}_t + \bar{\lambda} \left( V'_t - V_t \right) \text{ with boundary condition } V'_F = 0
\end{aligned}
$$

The target value-gradient is given by:

$$
\frac{\partial G'_t}{\partial t} = -\left( \frac{D\bar{r}}{D\vec{x}} \right)_t - \left( \frac{D\bar{f}}{D\vec{x}} \right)_t \left( G'_t \right) + \bar{\lambda} \left( G'_t - G_t \right) \tag{24}
$$

with boundary condition $G'_F = \vec{0}$, and where $\frac{D}{D\vec{x}} \equiv \frac{\partial}{\partial \vec{x}} + \frac{\partial \pi}{\partial \vec{x}} \frac{\partial}{\partial a}$. This is a DE that needs solving with equivalent care to which the trajectory was, and it may also be stiff. Note that in this equation, $\bar{r}$ is the full $\bar{r}$ as defined in Eq. 22. Also, in the case of an episodic problem where a final impulse of reward is given, an alternative boundary condition to this one may be required—see Appendix E.1 for a discussion and example.

For this policy, model and $\bar{\lambda} = 0$, the PGL weight update is:

$$
\Delta \vec{w} = \alpha \left( \frac{\partial R^\pi}{\partial \vec{w}} \right)_0 = \alpha \int_0^F \left( \frac{\partial G}{\partial \vec{w}} \right)_t \bar{\Omega}_t (G'_t - G_t) dt \tag{25}
$$

where $\bar{\Omega}_t = \frac{1-a_t^2}{c} \left( \frac{\partial \bar{f}}{\partial a} \right)_t^T \left( \frac{\partial \bar{f}}{\partial a} \right)_t$ and is positive semi-definite.

This integral is the exact equation for gradient ascent on $R^\pi$. Therefore, if implemented precisely, termination will occur at a constrained (with respect to $\vec{w}$) locally optimal trajectory (see Appendix A for optimality definitions). However, numerical methods are required to evaluate this integral and the other DEs in this section. For example, the above integral is most simply approximated as:

$$
\Delta \vec{w} = \alpha \sum_{t=0}^{F} \left( \frac{\partial G}{\partial \vec{w}} \right)_t \bar{\Omega}_t (G'_t - G_t) \Delta t
$$

which is very similar to the discrete-time case (Eq. 18).

The fact that this algorithm is gradient ascent means it can be speeded up with any of the fast optimisers available, e.g. RPROP (Riedmiller and Braun, 1993), which becomes very useful when $c$ is small and therefore $\Delta \vec{w}$ becomes very small.

Eq. 25 was derived for $\bar{\lambda} = 0$ although it can, in theory, be applied to other $\bar{\lambda}$. In this case, it is thought to be necessary to choose a full-rank version of $\bar{\Omega}$. However our results with bootstrapping with a function approximator are not good (see Section 4.5).

## 2.3 Residual Gradients

In this section we derive the formulae for full gradient descent on the value-gradient error function, according to Eq. 12 and Eq. 13. The particularly promising thing about this approach is that it has good convergence guarantees for any $\lambda$. This kind of full gradient descent method is known as *residual gradients* by Baird (1995) or as using the *Galerkinized equations* by Werbos (1998).

To calculate the total derivative of $E$, it is easier to first write $E$ in recursive form:

$$E(\vec{x}_t, \vec{w}) = \frac{1}{2}(G_t - G'_t)^T \Omega_t (G_t - G'_t) + E(\vec{x}_{t+1}, \vec{w})$$

with $E(\vec{x}_F, \vec{w}) = 0$ at the terminal time step. This gives a total derivative:

$$\left(\frac{\partial E}{\partial \vec{w}}\right)_t = \left(\left(\frac{\partial G}{\partial \vec{w}}\right)_t - \left(\frac{\partial G'}{\partial \vec{w}}\right)_t\right)\Omega_t(G_t - G'_t) + \left(\frac{\partial \pi}{\partial \vec{w}}\right)_t\left(\frac{\partial f}{\partial a}\right)_t\left(\frac{\partial E}{\partial \vec{x}}\right)_{t+1} + \left(\frac{\partial E}{\partial \vec{w}}\right)_{t+1}$$

with $\left(\frac{\partial E}{\partial \vec{w}}\right)_F = \vec{0}$ and where $\left(\frac{\partial E}{\partial \vec{x}}\right)_{t+1}$ is found recursively by

$$\left(\frac{\partial E}{\partial \vec{x}}\right)_t = \left(\left(\frac{\partial G}{\partial \vec{x}}\right)_t - \left(\frac{\partial G'}{\partial \vec{x}}\right)_t\right)\Omega_t(G_t - G'_t) + \left(\left(\frac{\partial \pi}{\partial \vec{x}}\right)_t\left(\frac{\partial f}{\partial a}\right)_t + \left(\frac{\partial f}{\partial \vec{x}}\right)_t\right)\left(\frac{\partial E}{\partial \vec{x}}\right)_{t+1}$$

with $\left(\frac{\partial E}{\partial \vec{x}}\right)_F = \vec{0}$. Note that this goes further than doing residual-gradients for VL in that there is a consideration for $\frac{\partial E}{\partial \vec{x}}$. This is necessary for true gradient-descent on $E$ with respect to the weights, since in this paper we say the value function and greedy policy are tightly coupled, and therefore updating $\vec{w}$ will immediately change the trajectory. This can be verified by evaluating $\frac{\partial E}{\partial \vec{w}}$ numerically.

Although this weight update performs relatively well in the experiments of Section 4, our general experience of this algorithm is that it often gets stuck in far-from-optimal local minima. This was quite puzzling and was not explained by previous criticisms of residual gradients (for example, see Baird, 1995), since these only applied to stochastic scenarios. It seems that many of the local minima of $E$ are not necessarily valid local maxima of $R^\pi$. We speculate that choosing to include the residual gradient terms is analogous to choosing to maximise a function $f(x)$ by gradient descent on $(f'(x))^2$. This makes it difficult to distinguish the maxima from the unwanted minima, inflections and saddle points, and although the situations are not identical, an effect like this may be reducing the effectiveness of the residual gradients weight update. This contrasts with the non-residual gradients approach (Eq. 14) where $\Delta \vec{w} = \alpha \frac{\partial R}{\partial \vec{w}}$ which is analogous to maximising $f(x)$ directly (if $\lambda = 1$; see Section 2.1). By the arguments of Section 1.4, we would expect this explanation to apply to VL with residual gradients too.

To illustrate this problem by a specific example, consider a variant of the 1-step Toy Problem with $k = 0$, modified so that the final reward is $R(x_1) = -x_1^2 + 4\cos(x_1)$ instead of the usual $R(x_1) = -x_1^2$. Then the optimal policy is $\pi^*(x_0) = -x_0$. Let the function approximator be $V(x_1, w) = -x_1{}^2 + wx_1$, so that the greedy policy with this model gives $\pi(x_0, w) = w/2 - x_0$. This gives $R^\pi(x_0, w) = -w^2/4 + 4\cos(w/2)$ which has just one maximum, at $w = 0$, corresponding to the optimal policy. The residual value-gradient error is $E(x_0, w) = \frac{1}{2}(G_1 - G'_1)^2 = \frac{1}{2}(w - 4\sin(w/2))^2$ which has many local minima (see Figure 5), only one of which corresponds to the optimal policy. The spurious minima in $E$ correspond to points of inflection in $R_0^\pi$, of which there are infinitely many. Therefore gradient descent on $E$ is more likely than not to converge to an incorrect solution, whereas gradient ascent on $R^\pi$ will converge to the correct solution.

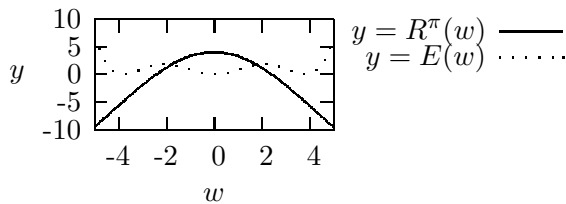

Figure 5: Graphs showing the spurious minima traps that can exist for Residual Gradient methods compared to direct methods.

## 3. Actor-Critic Architectures

This section discusses the use of actor-critic architectures with VGL. It shows that in some circumstances the actor-critic architecture can be shown to be equivalent to a simpler architecture. While this can be used to provide convergence guarantees for the actor-critic architecture, it also makes the actor-critic architecture redundant in these circumstances.

An actor-critic architecture (Barto et al., 1983; Werbos, 1998; Konda and Tsitsiklis, 2003) uses two neural-networks, or more generally two function approximators, in a control problem. The first neural-network, parametrised by weight vector $\vec{z}$, is the "actor" which provides the policy, $\pi(\vec{x}, \vec{z})$. In an actor-critic architecture the greedy policy is not used, since the actor neural-network is the policy. The second neural-network, parametrised by a weight vector $\vec{w}$, is the "critic" and provides the value function $V(\vec{x}, \vec{w})$.

For this section we extend the definition of $V'$ and $G'$ to apply to trajectories found by policies other than the greedy policy. To define $V'$ for an arbitrary policy $\pi(\vec{x}, \vec{z})$ and value function $V(\vec{x}, \vec{w})$ we use:

$$V'(\vec{x}, \vec{w}, \vec{z}) = r(\vec{x}, \pi(\vec{x}, \vec{z})) + \lambda V'(f(\vec{x}, \pi(\vec{x}, \vec{z})), \vec{w}, \vec{z}) + (1 - \lambda)V(f(\vec{x}, \pi(\vec{x}, \vec{z})), \vec{w}) \quad (26)$$

with $V'(\vec{x}_F, \vec{w}, \vec{z}) = 0$. This gives $G'(\vec{x}, \vec{w}, \vec{z}) = \frac{\partial V'(\vec{x}, \vec{w}, \vec{z})}{\partial \vec{x}}$ and for a given trajectory we can use the shorthand $V'_t = V'(\vec{x}_t, \vec{w}, \vec{z})$ and $G'_t = G'(\vec{x}_t, \vec{w}, \vec{z})$.

The value-gradient version of the actor training equation is as follows (Werbos, 1998, Eq. 11):

$$\Delta \vec{z} = \alpha \sum_{t \geq 0} \left(\frac{\partial \pi}{\partial \vec{z}}\right)_t \left(\left(\frac{\partial r}{\partial a}\right)_t + \left(\frac{\partial f}{\partial a}\right)_t G_{t+1}\right) \quad (27)$$

where $\alpha$ is a small positive learning-rate and $G_{t+1}$ is calculated by the critic. This equation is almost identical to the PGL equation (BPTT, Eq. 15) except that $G_{t+1}$ has been substituted for $\left(\frac{\partial R^\pi}{\partial \vec{x}}\right)_{t+1}$. Eq. 27 is a non-standard actor training equation, however in Appendix F we prove it is equivalent to at least one other actor training equation and demonstrate how it automatically incorporates exploration.

The critic training equation would attempt to attain $G_t = G'_t$ for all $t$ by some appropriate method, for example by Eq. 14, which is the only update we consider here.

In this section, it is useful to define the notion of a hypothetical *ideal function approximator*. This is a function approximator that can be assumed to have enough degrees of freedom, and a strong enough learning algorithm, to attain its desired objectives exactly, e.g. $G_t = G'_t$ for all $t$, exactly. We also refer to an ideal critic and an ideal actor, which are based on ideal function approximators.

Training an actor and a critic together gives several possibilities of implementation; either one could be fixed while training the other, or they could both be trained at the same time. We only analyse the situations of keeping one fixed while training the other. Doing a long iterative process (training one) within another long iterative process (training the other) is very bad for efficiency, which may make the cases analysed seem infeasible and therefore of little relevance. However, the analyses below show both of these situations have equivalent architectures that are efficient and feasible to implement, and therefore are relevant.

It is noted that the results in this section should also apply to VL actor-critic system, since as discussed in Section 1.4, VL is a stochastic approximation to VGL.

## 3.1 Keeping the Actor Fixed while training the Critic

In this scenario we keep the actor fixed while training an ideal critic fully to convergence, and then apply one iteration of actor training, and then repeat until both have converged. It is shown that, if the critic is ideal, then this scenario is equivalent to back-propagation through time (BPTT) for any $\lambda$.

Since the actor is fixed, the trajectory is fixed; therefore an ideal critic will be able to attain the objective of $G_t = G'_t$ for all $t$ along the trajectory. Then since $G_t = G'_t$ for all $t$, we have $G_t = \left(\frac{\partial R^\pi}{\partial \vec{x}}\right)_t$ for all $t$ (proof is in Appendix A, lemma 5). Therefore the actor's weight update equation (Eq. 27) becomes identical to Eq. 15. Therefore we could omit the critic and replace $G$ with $\frac{\partial R^\pi}{\partial \vec{x}}$ in the actor training equation (i.e. we remove the inner iterative process, and remove the actor-critic architecture), and we are left with BPTT. This shows that this actor-critic is guaranteed to converge, since it is the same as BPTT.

The above argument assumed the critic was ideal. This may be an unrealistic assumption since a real function approximator can have only finite flexibility. However, the objective of the function approximator is to learn $G_t = G'_t$ for all $t$, and this goal *can* be achieved exactly; simply by removing the critic. In effect, by removing the critic, a *virtual* ideal function approximator is obtained. It is assumed that there would be no advantage in using a non-ideal critic.

Conclusion: BPTT is equivalent to the idealised version of this architecture, and therefore the idealised version of this architecture is guaranteed to converge to a constrained locally optimal trajectory (see Appendix A for optimality definitions). The idealised version of this architecture is attainable by removing the critic.

## 3.2 Keeping the Critic Fixed while training the Actor

In this scenario we consider keeping the critic fixed while training an ideal actor fully to convergence, and then apply one iteration of training to the critic and repeat.

The actor weight update equation (Eq. 27) tries to maximise $Q(\vec{x}_t, a_t, \vec{w})$ with respect to $a_t$ at each time step $t$. If this is fully achieved, then the greedy policy will be satisfied. Therefore the ideal actor can be removed and replaced by the greedy policy, to get the same algorithm. This again removes the innermost iterative step, and removes the actor-critic architecture. Again, it is assumed that there would be no advantage in using a non-ideal actor.

Now when it comes to the critic-training step (Eq. 14), do we allow for the fact that changing the critic weights is going to change the actor in a predictable way? If so, then we treat $\Omega_t$ as defined in Eq. 19. Otherwise we are working as if the actor and critic are fully separated, and we are free to choose $\Omega_t$. Having made this choice and substitution, the actor is redundant.

Conclusion: Keeping the critic fixed while training an idealised actor is equivalent to using just the critic with a greedy policy. This idealised architecture can be efficiently attained by removing the actor.

## 4. Experiments

In this section a comparison is made between the performance of several weight update strategies on various tasks. The weight update formulae considered are summarised in Table 1.

The first few experiments are all based on the *n-step Toy Problem* defined in Section 1.1.1 The choice of this problem domain was made because it is smooth, deterministic, concave and possible to make the experiments easily describable and reproducible. Within this choice, it makes a level playing field for comparison between VL and VGL algorithms.

The final experiment is a neural-network based experiment on a Lunar-Lander problem specified in Appendix E. This is a new benchmark problem defined for this paper; the problem with most current existing benchmark problems was that they are only defined for continuing tasks, discrete state-space tasks or tasks not well suited to local exploration.

In the Toy Problem experiments, all weight components were initialised with a uniform random distribution over a range from $-10$ to $+10$. The experiments were based on 1000 trials. The stopping criteria used for each trial were as follows: A trial was considered a success when $|w_i - w_i^*| < 10^{-7}$ for all components $i$. A trial was considered a failure if any component of $\vec{w}$ became too large for the computer's accuracy, or when the number of iterations in any trial exceeded $10^7$. A function $RND(\epsilon)$ is defined to return a normally distributed random variable with mean 0 and standard deviation $\epsilon$. The variables $c_1$, $c_2$, $c_3$ are real constants specific to each experiment designed to allow further variation in the problem specifications.

| Weight Update Formula | Abbreviation |
|---|---|
| Value-Learning Update, TD($\lambda$), (Eq. 10). | VL($\lambda$) |
| Value-Gradient Learning Update (Eq. 18), using full $\Omega_t$ matrix from Eq. 19. | VGL$\Omega(\lambda)$ |
| Value-Gradient Learning Update (Eq. 14), but using $\Omega_t$ as the identity matrix for all $t$. | VGL($\lambda$) |
| Value-Gradient Learning Update, residual gradients (Eq. 13), but using $\Omega_t$ as the identity matrix for all $t$. | VGLRG($\lambda$) |

Table 1: Weight update formulae considered.

## 4.1 Experiment 1: One-step Toy Problem

In this experiment the one-step Toy Problem with $k = 0$ was considered from a fixed start point of $x_0 = 0$. A function approximator for $V$ with just two parameters $(w_1, w_2)$ was used, and defined separately[4] for the two time steps:

$$V(x_t, w_1, w_2) = \begin{cases} -(x_1 - c_1)^2 + w_1 x_1 + w_2 & \text{if } t = 1 \\ 0 & \text{if } t = 2 \end{cases}$$

where $c_1$ is a real constant.

Using this model and function approximator definition, it is possible to calculate the functions $Q(x, a, \vec{w})$, $G(x, \vec{w})$ and the $\epsilon$-greedy policy $\pi(x, \vec{w})$ (which again must be defined differently for each time step). These functions are listed in the left-hand column of Table 2. Using these formulae, and the model functions again, the full trajectory is calculated in the right-hand column of Table 2. Also in this right-hand column, $V'$, $G'$ and $\Omega$ have been calculated for each time step using Eq. 4, Eq. 6 and Eq. 19 respectively. These formulae would have to be evaluated sequentially from top to bottom.

The $\epsilon$-greedy policy was necessary for the VL experiments. When applying the weight update formulae, expressions for $\frac{\partial V}{\partial \vec{w}}$ and $\frac{\partial G}{\partial \vec{w}}$ were calculated analytically from the functions given in the left column of Table 2. For example, for this function approximator we find $\left(\frac{\partial V}{\partial w_1}\right)_1 = x_1$, $\left(\frac{\partial V}{\partial w_2}\right)_1 = 1$ and $\left(\frac{\partial G}{\partial w_1}\right)_1 = 1$, etc.

Note that as $w_2$ does not affect the trajectory, this component was not used as part of the stopping criteria.

Results for these experiments using the VL($\lambda$) and VGL($\lambda$) algorithms are shown in Table 3. This set of experiments verifies the failure of VL when exploration is removed; the slowing down of VL when it is too low; and the blowing-up of VL when it is too high (in this case failure tended to occur because the size of weights exceeded the computer's range). The efficiency and success rate of the VGL experiments is much better than for the VL experiments, and this is true for both values of $c_1$ tested. In fact, the problem is trivially easy for the VGL($\lambda$) algorithm, but causes the VL($\lambda$) algorithm considerable problems.

To gain some further insight into the different behaviour of these two algorithms we can look at the differential equations that the weights obey. For the VGL($\lambda$) system of this experiment, by going through the equations of the right-hand column of Table 2 and the VGL($\lambda$) weight update equations, we can eliminate all variables except for the weights and constants to obtain a self-contained pair of weight update equations:

$$\begin{cases} \Delta w_1 = -\alpha \left(2c_1 + w_1\right) \\ \Delta w_2 = 0 \end{cases}$$

---

4. See Section 1.1.1 for an explanation on this abuse of notation.

| Function approximator and $\epsilon$-Greedy Policy | Sequential trajectory equations |
|---|---|
| Time step 1:<br>$V(x_1, w_1, w_2) = -(x_1 - c_1)^2 + w_1 x_1 + w_2$<br>$\Rightarrow \quad G(x_1, w_1, w_2) = 2(c_1 - x_1) + w_1$<br>$\quad\quad Q(x_0, a_0, \vec{w}) = -(x_0 + a_0 - c_1)^2$<br>$\quad\quad\quad\quad + w_1(x_0 + a_0) + w_2$<br>$\quad\quad \pi(x_0, \vec{w}) = (2c_1 - 2x_0 + w_1)/2 + RND(\epsilon)$<br>Time step 2:<br>$V(x_2, w_1, w_2) = 0$<br>$\Rightarrow \quad G(x_2, w_1, w_2) = 0$ | $x_0 \leftarrow 0$<br>$a_0 \leftarrow (2c_1 - 2x_0 + w_1)/2 + RND(\epsilon)$<br>$x_1 \leftarrow x_0 + a_0$<br>$V'_1 \leftarrow -x_1{}^2$<br>$G'_1 \leftarrow -2x_1$<br>$V_1 \leftarrow -(x_1 - c_1)^2 + x_1 w_1 + w_2$<br>$G_1 \leftarrow 2(c_1 - x_1) + w_1$<br>$\Omega_0 \leftarrow \frac{1}{2}$ |
| Optimal Policy: $\pi^*(x_0) = -x_0$ | |
| Optimal Weights: $w_1{}^* = -2c_1$ | |

Table 2: Functions and Trajectory Variables for Experiment 1.

| $c_1$ | $\epsilon$ | $\alpha = 0.01$ | | | $\alpha = 0.1$ | | | $\alpha = 1.0$ | | |
|---|---|---|---|---|---|---|---|---|---|---|
| | | Success rate | Iterations (Mean) | (s.d.) | Success rate | Iterations (Mean) | (s.d.) | Success rate | Iterations (Mean) | (s.d.) |
| Results for algorithm VL($\lambda$) | | | | | | | | | | |
| 0 | 10 | 66.4% | 1075.1 | 293.35 | 0.0% | | | 0.0% | | |
| 0 | 1 | 100.0% | 1715.8 | 343.31 | 87.6% | 163.52 | 31.948 | 3.8% | 134.86 | 59.643 |
| 0 | 0.1 | 100.0% | 172445 | 31007 | 89.5% | 17160 | 3033.6 | 16.5% | 1527.6 | 118.39 |
| 0 | 0 | 0.0% | | | 0.0% | | | 0.0% | | |
| 10 | 1 | 99.4% | 6048.5 | 270.86 | 0.0% | | | 0.0% | | |
| Results for algorithm VGL($\lambda$) | | | | | | | | | | |
| 0 | 0 | 100.0% | 1728.2 | 112.05 | 100.0% | 166.15 | 11.481 | 100.0% | 1 | 0 |
| 10 | 0 | 100.0% | 1898.5 | 51.986 | 100.0% | 181.59 | 14.861 | 100.0% | 1 | 0 |

Table 3: Results for Experiment 1. Note because this is a 1-step problem, $\lambda$ is irrelevant.

Taking $\alpha$ to be sufficiently small, these become a pair of coupled differential equations. The solution is a straight line across the $\vec{w}$ plane directly to the optimal solution $w_1^* = -2c_1$.

Doing the same for the VL($\lambda$) system, and integrating over the random variable $RND(\epsilon)$ to average out the effects of exploration, gives a similar pair of coupled weight update equations:

$$\begin{cases} \langle \Delta w_1 \rangle = -\alpha \left(c_1 + \frac{w_1}{2}\right) \left(2\epsilon^2 + c_1{}^2 + 2c_1 w_1 + \frac{w_1{}^2}{2} + w_2\right) \\ \langle \Delta w_2 \rangle = -\alpha \left(c_1{}^2 + 2c_1 w_1 + \frac{w_1{}^2}{2} + w_2\right) \end{cases}$$

There is no known full analytical solution to this pair of equations. However it is clear that the second equation is continually aiming to achieve $w_2 = -\left(c_1{}^2 + 2c_1 w_1 + w_1{}^2/2\right)$. In the case that this is achieved, both equations would then simplify to the VGL coupled equations, but with a magnitude proportional to $\epsilon^2$. This shows that if $\epsilon = 0$, the value-gradient part of these equations vanishes. It is also noted that in this case experiments show learning fails. Hence it is speculated that none of the other terms in the VL($\lambda$) coupled equations are doing anything beneficial, and that it is unlikely they will ever do so even in more complicated systems. Very informally, this example illustrates how VGL applies just the "important bits" of a VL weight update (in this example at least).

## 4.2 Experiment 2: Two-step Toy Problem, with Sufficiently Flexible Function Approximator

In this experiment the two-step Toy Problem with $k = 1$ is considered from a fixed start point of $x_0 = 0$. A function approximator is defined differently at each time step, by four weights in total:

$$V(x_t, w_1, w_2, w_3, w_4) = \begin{cases} -c_1 x_1{}^2 + w_1 x_1 + w_2 & \text{if } t = 1 \\ -c_2 x_2{}^2 + w_3 x_2 + w_4 & \text{if } t = 2 \\ 0 & \text{if } t = 3 \end{cases}$$

where $c_1$ and $c_2$ are real positive constants. The consequential functions and variables for this experiment are found and presented in a similar manner as for Experiment 1, in Table 4.

For ease of implementation of the residual-gradients algorithm, the expressions in the right-hand column of Table 4 for $G_t$ and $G'_t$ were used to implement a sum-of-squares error function $E(\vec{w})$ (Eq. 12), with $\Omega_t = 1$. Numerical differentiation on this function was then used to implement the gradient descent. For a larger scale system, it would be more efficient and accurate to use the recursive equations given in Section 2.3.

Results for the experiments are given in Table 5. These results show all VGL experiments performing significantly better than the corresponding VL experiments; in most cases by around two orders of magnitude. The results also show that for all of the VGL($\lambda$) algorithm results, increasing $\alpha$ from 0.01 to 0.1 brings the number of iterations down by a factor of approximately 10, which hints that further efficiency of the VGL algorithms could be attained.

| Function approximator and $\epsilon$-Greedy Policy | Sequential trajectory equations |
|---|---|
| **Time step 1:** $V(x_1, w_1, w_2) = -c_1 x_1{}^2 + w_1 x_1 + w_2$ $\Rightarrow \ G(x_1, w_1, w_2) = -2c_1 x_1 + w_1$ $\qquad Q(x_0, a_0, \vec{w}) = -k a_0{}^2 - c_1(x_0 + a_0)^2$ $\qquad\qquad + w_1(x_0 + a_0) + w_2$ $\qquad \pi(x_0, \vec{w}) = \frac{w_1 - 2c_1 x_0}{2(c_1 + k)} + RND(\epsilon)$ **Time step 2:** $V(x_2, w_3, w_4) = -c_2 x_2{}^2 + w_3 x_2 + w_4$ $\Rightarrow \ G(x_2, w_3, w_4) = -2c_2 x_2 + w_3$ $\qquad Q(x_1, a_1, \vec{w}) = -k a_1{}^2 - c_2(x_1 + a_1)^2$ $\qquad\qquad + w_3(x_1 + a_1) + w_4$ $\qquad \pi(x_1, \vec{w}) = \frac{w_3 - 2c_2 x_1}{2(c_2 + k)} + RND(\epsilon)$ **Time step 3:** $V(x_3) = 0$ $\Rightarrow \ G(x_3) = 0$ | $x_0 \leftarrow 0$ $a_0 \leftarrow \frac{w_1 - 2c_1 x_0}{2(c_1 + k)} + RND(\epsilon)$ $x_1 \leftarrow x_0 + a_0$ $a_1 \leftarrow \frac{w_3 - 2c_2 x_1}{2(c_2 + k)} + RND(\epsilon)$ $x_2 \leftarrow x_1 + a_1$ $V'_2 \leftarrow -x_2{}^2$ $G'_2 \leftarrow -2x_2$ $V_2 \leftarrow -c_2 x_2{}^2 + w_3 x_2 + w_4$ $G_2 \leftarrow -2c_2 x_2 + w_3$ $\Omega_1 \leftarrow \frac{1}{2(c_2 + k)}$ $V'_1 \leftarrow -k a_1{}^2 + \lambda V'_2 + (1 - \lambda)V_2$ $G'_1 \leftarrow \frac{2c_2 k a_1 + k(\lambda G'_2 + (1 - \lambda)G_2)}{c_2 + k}$ $V_1 \leftarrow -c_1 x_1{}^2 + w_1 x_1 + w_2$ $G_1 \leftarrow -2c_1 x_1 + w_1$ $\Omega_0 \leftarrow \frac{1}{2(c_1 + k)}$ |
| **Optimal Weights:** $w_1{}^* = w_3{}^* = 0$ | |

Table 4: Functions and Trajectory Variables for Experiment 2.

The optimal value function, denoted by $V^*$, for this experiment is

$$V^*(x_t) = \begin{cases} -\frac{k}{1+k} x_1{}^2 & \text{if } t = 1 \\ -x_2{}^2 & \text{if } t = 2 \end{cases}$$

For this reason most experiments were done with $c_1 = \frac{1}{2}$ and $c_2 = 1$. However the only necessity is to have $c_1 > 0$ and $c_2 > 0$, since these are required to make the greedy policy produce continuous actions; a problematic issue for all value function architectures.

### 4.3 Experiment 3: Divergence of Algorithms With Two-step Toy Problem

We now study the Toy Problem to try to find a set of parameters that cause learning to become unstable. Surprisingly the two-step Toy Problem is sufficiently complex to provide examples of divergence, both with and without bootstrapping. By the principles argued in Section 1.4, we would expect these examples that were found for VGL to also cause divergence with the corresponding VL methods. This is confirmed empirically.

If we take the previous experiment and consider the VGL$\Omega(\lambda)$ weight update then the only two weights that change are $\vec{w} = (w_1, w_3)^T$. The weight update equation for these two weights can be found analytically by substituting all the equations of the right hand side of

| Weight Update Algorithm ($\lambda$) | $\epsilon$ | $c_1$ | $c_2$ | $\alpha = 0.01$ | | | $\alpha = 0.1$ | | |
|---|---|---|---|---|---|---|---|---|---|
| | | | | Success rate | Iterations | | Success rate | Iterations | |
| | | | | | (Mean) | (s.d.) | | (Mean) | (s.d.) |
| VL(1) | 1 | 0.5 | 1 | 100.0% | 244122 | 252234 | 91.3% | 736030 | 743920 |
| VL(1) | 0.1 | 0.5 | 1 | 100.0% | 135588 | 17360 | 100.0% | 21406.6 | 8641 |
| VGL(1) | 0 | 0.5 | 1 | 100.0% | 1596.8 | 72.58 | 100.0% | 152.5 | 13.79 |
| VGL$\Omega$(1) | 0 | 0.5 | 1 | 100.0% | 6089.1 | 340.09 | 100.0% | 600.7 | 47.06 |
| VGLRG(1) | 0 | 0.5 | 1 | 100.0% | 794.6 | 40.30 | 100.0% | 72.2 | 5.36 |
| VL(0) | 1 | 0.5 | 1 | 100.0% | 244368 | 252114 | 91.6% | 734029 | 742977 |
| VL(0) | 0.1 | 0.5 | 1 | 100.0% | 138073 | 17630 | 99.9% | 21918 | 8664 |
| VGL(0) | 0 | 0.5 | 1 | 100.0% | 1743.7 | 103.41 | 100.0% | 166.2 | 12.81 |
| VGL$\Omega$(0) | 0 | 0.5 | 1 | 100.0% | 6516.5 | 375.99 | 100.0% | 643.0 | 39.12 |
| VGLRG(0) | 0 | 0.5 | 1 | 100.0% | 1252.4 | 92.81 | 100.0% | 118.2 | 11.63 |
| VL(1) | 0.1 | 4 | 1 | 100.0% | 228336 | 60829 | 100.0% | 78364 | 62085 |
| VGL(1) | 0 | 4 | 1 | 100.0% | 5034.7 | 340.6 | 100.0% | 495.1 | 35.7 |
| VL(1) | 0.1 | 0.1 | 1 | 100.0% | 134443 | 16614 | 100.0% | 20974 | 8569 |
| VGL(1) | 0 | 0.1 | 1 | 100.0% | 1516.2 | 89.5 | 100.0% | 144.4 | 13.8 |

Table 5: Results for various algorithms on Experiment 2.

Table 4 into the VGL$\Omega$($\lambda$) weight update equation, and using $\epsilon = 0$, giving:

$$\Delta \vec{w} = \alpha D E D \vec{w}$$

with $E = -2 \begin{pmatrix} (k + \lambda(1+b)(b(k+1)+1) - bk) & (\lambda(k+1)(b+1) - k) \\ 1 + b(k+1) & (k+1) \end{pmatrix}$, $b = \left(\frac{\partial \pi}{\partial x}\right)_1 = \frac{-c_2}{c_2 + k}$ and $D = \begin{pmatrix} 1/2(k + c_1) & 0 \\ 0 & 1/2(k + c_2) \end{pmatrix}$

We can consider more types of function approximator by defining the weight vector $\vec{w}$ to be linear system of two new weights $\vec{p} = (p_1, p_2)^T$ such that $\vec{w} = F\vec{p}$ and where $F$ is a $2 \times 2$ constant real matrix. If the VGL$\Omega$($\lambda$) weight update equation is now recalculated for these new weights then the dynamic system for $\vec{p}$ is:

$$\Delta \vec{p} = \alpha (F^T D E D F) \vec{p} \qquad (28)$$

Taking $\alpha$ to be sufficiently small, then the weight vector $\vec{p}$ evolves according to a continuous-time linear dynamic system, and this system is stable if and only if the matrix product $F^T D E D F$ is stable (i.e. if the real part of every eigenvalue of this matrix product is negative).

The VGL($\lambda$) system weight update can also be derived and that system is identical to Eq. 28 but with the leftmost $D$ matrix omitted.

Choosing $\lambda = 0$, with $c_1 = c_2 = k = 0.01$ and $F = D^{-1} \begin{pmatrix} 10 & 1 \\ -1 & -1 \end{pmatrix}$ leads to divergence for both the VGL($\lambda$) and VGL$\Omega$($\lambda$) systems. Empirically, we found that these parameters cause the VL(0) algorithm to diverge too. This is a specific counterexample for the VL system which is "on-policy" and equivalent to Sarsa. Previous examples of divergence for a function approximator with bootstrapping have usually been for "off-policy" learning (see for example, Baird, 1995; Tsitsiklis and Roy, 1996b). Tsitsiklis and Roy (1996a) describe an "on-policy" counterexample for a non-linear function approximator, but this is not for the greedy policy.

| Algorithm ($\lambda$) | Divergence example found | Proven to converge |
|---|---|---|
| VL(1) | Yes | No |
| VL(0) | Yes | No |
| VGL(1) | Yes | No |
| VGL(0) | Yes | No |
| VGL$\Omega$(1) | No | Yes |
| VGL$\Omega$(0) | Yes | No |

Table 6: Results for Experiment 3: Which algorithms can be made to diverge?

Also, perhaps surprisingly, it is possible to get instability with $\lambda = 1$ with the VGL($\lambda$) system. Substituting $c_2 = k = 0.01$, $c_1 = 0.99$ and $F = D^{-1} \begin{pmatrix} -1 & -1 \\ 10 & 1 \end{pmatrix}$ makes the VGL(1) system diverge. This result has been empirically verified to carry over to the VL(1) system too, i.e. this is a result where Sarsa(1) and TD(1) diverge. This highlights the difficulty of control-problems in comparison to prediction tasks. A prediction task is easier, since as Sutton (1988) showed, the $\lambda = 1$ system is equivalent to gradient descent on the sum-of-squares error $E = \Sigma_t (V'_t - V_t)^2$, and so convergence is guaranteed for a prediction task. However in a control problem, even when there is no bootstrapping, changing one value of $V'_t$ affects the others by altering the greedy actions. This problem is resolved by using the VGL$\Omega$(1) weight update.

The results of this section are summarised in Table 6.

## 4.4 Experiment 4: Two-step Toy Problem, with Insufficiently Flexible Function Approximator

In this experiment the two-step Toy Problem with $k = 2$ was considered from a fixed start point of $x_0 = 0$. A function approximator with just one weight component, ($w_1$), was defined differently at each time step:

$$V(x_t, w_1) = \begin{cases} -c_1 x_1{}^2 + w_1 x_1 & \text{if } t = 1 \\ -c_2 x_2{}^2 + (w_1 - c_3) x_2 & \text{if } t = 2 \\ 0 & \text{if } t = 3 \end{cases}$$

Here $c_1 = 2$, $c_2 = 0.1$ and $c_3 = 10$ are constants. These were designed to create some conflict for the function approximator's requirements at each time step. The optimal actions are $a_0 = a_1 = 0$, and therefore the function approximator would only be optimal if it could achieve $\frac{\partial V(x, \vec{w})}{\partial x} = 0$ at $x = 0$ for both time steps. The presence of the $c_3$ term makes this impossible, so a compromise must be made.

| Sequential equations | |
|---|---|
| $x_0 \leftarrow 0$ | $G'_2 \leftarrow -2x_2$ |
| $a_0 \leftarrow \frac{w_1 - 2c_1 x_0}{2(c_1 + k)}$ | $G_2 \leftarrow -2c_2 x_2 + w_1$ |
| $x_1 \leftarrow x_0 + a_0$ | $G'_1 \leftarrow \frac{2c_2 k a_1 + k(\lambda G'_2 + (1-\lambda)G_2)}{c_2 + k}$ |
| $a_1 \leftarrow \frac{w_1 - c_3 - 2c_2 x_1}{2(c_2 + k)}$ | $G_1 \leftarrow -2c_1 x_1 + w_1$ |
| $x_2 \leftarrow x_1 + a_1$ | $\Omega_1 \leftarrow \frac{1}{2(c_2 + k)}$ |
| $\Omega_0 \leftarrow \frac{1}{2(c_1 + k)}$ | |

Table 7: Trajectory Variables for Experiment 4.

Only the value-gradient algorithms were considered in this experiment; hence there was no need for exploration or use of the $\epsilon$-greedy policy. The equations for the trajectory are very similar as to Experiment 2, and so only the key results are listed in Table 7. A different stopping criterion was used in this experiment, since each algorithm converges to a different fixed point. The stopping condition used was $|\Delta w_1| < (10^{-7}\alpha)$. Once the fixed point had been reached we noted the value of $R$, the total reward for that trajectory, in Table 8. Each algorithm experiment used $\alpha = 0.01$, attained 100% convergence, and produced the same value of $R$ each run. In these trials, $\alpha$ was not varied, and the iteration result columns are not very meaningful, since it can be shown for each algorithm that $\Delta w_1$ is a linear function of $w_1$. This indicates that any of the algorithms could be made arbitrarily fast by fine tuning $\alpha$, and also confirms that there is only one fixed point for each algorithm.

The different values of $R$ in the results show that the different algorithms have varying degrees of optimality with respect to $R$. The first algorithm (VGL$\Omega(1)$) is the only one that is really optimal with respect to $R$ (subject to the constraints imposed by the awkward choice of function approximator), since it is equivalent to gradient ascent on $R^\pi$, as shown in section 2.1.

It is interesting that the other algorithms converge to different suboptimal points, compared to VGL$\Omega(1)$. This shows that introducing the $\Omega_t$ terms and using $\lambda = 1$ balances the priorities between minimising $(G'_1 - G_1)^2$ and minimising $(G'_2 - G_2)^2$ appropriately so as to finish with an optimal value for $R$. Different values for the constants $c_1$ and $c_2$ were chosen to emphasise this point. The relative rankings of the other algorithms may change in other problem-domains, but it is expected that VGL$\Omega(1)$ would always be at the top. Hence we use this algorithm as our standard choice of algorithm out of those listed in this paper; to be used in conjunction with a robust optimiser, e.g. RPROP.

## 4.5 Experiment 5: One-Dimensional Lunar-Lander Problem

The objective of this experiment was to learn the "Lunar-Lander" problem described in Appendix E. The value-function was provided by a fully connected multi-layer perceptron (MLP) (see Bishop (1995) for details). The MLP had 3 inputs, one hidden layer of 6 units, and one output in the final layer. Additional shortcut connections connected all input units directly to the output layer. The activation functions were standard sigmoid functions in the input and hidden layers, and an identity function (i.e. linear activation) in the output unit.

| Weight Update Algorithm ($\lambda$) | $R$ | Iterations | |
|---|---|---|---|
| | | (Mean) | (s.d.) |
| VGL$\Omega$(1) | $-2.65816$ | 2327 | 247 |
| VGLRG(1) | $-2.68083$ | 183 | 19 |
| VGL(1) | $-2.79905$ | 500 | 51 |
| VGL$\Omega$(0) | $-2.82344$ | 3532 | 352 |
| VGLRG(0) | $-3.97316$ | 256 | 21 |
| VGL(0) | $-5.76701$ | 1077 | 87 |

Table 8: Results for Experiment 4, ranked by $R$.

The input to the neural-network was $(h/100, v/10, u/50)^T$, and the output was multiplied by 100 to give the value function. Each weight in the neural-network was initially randomised to lie in $[-1, 1]$, with uniform probability distribution.

In this section results are presented as pairs as diagrams. The left-hand diagrams show a graph of the total reward for all trajectories versus the number of training iterations, and compare performance to that of an optimal policy. The optimal policy's performance was calculated by the theory described in Appendix E.2. That appendix also shows example optimal trajectories. The right-hand diagrams show a cross-section through state space, with the y-axis showing height and the x-axis showing velocity. The final trajectories obtained are shown as curves starting at a diamond symbol and finishing at $h = 0$.

All algorithms used in this section were the continuous-time counterparts to those stated, and are described in Section 2.2. Also, all weight updates were combined with RPROP for acceleration. For implementing RPROP, the weight update for all trajectories was first accumulated, and then the resulting weight update was fed into RPROP at the end of each iteration. RPROP was used with the default parameters defined by Riedmiller and Braun (1993).

Results for the task of learning one trajectory from a fixed start point are shown in Figure 6 for the VGL$\Omega$ algorithm. The results were averaged over 10 runs. VGL$\Omega$ worked better on this task than VGL. VGL sometimes produced unstable or suboptimal solutions. However, both could manage the task well with $c = 1$. It was not possible to get VL to work on this task at all. VL failed with a greedy policy, as expected, as there is no exploration. However VL also failed on this task when using the $\epsilon$-greedy policy. We suspect the reason for this was that random exploration was producing more unwanted noise than useful information; random exploration makes the spacecraft fly the wrong way and learn the wrong values. We believe this makes a strong case for using value-gradients and a deterministic system instead of VL with random exploration.

The kink in the left diagram of Figure 6 was caused because $c$ was low, so the gradient $\frac{\partial R^\pi}{\partial \vec{w}}$ was tiny for the first few iterations. It seemed that RPROP would cause the weight change to build up momentum quickly and then temporarily overshoot its target before bringing it back under control. It was very difficult to learn this task with such a small $c$ value without RPROP.

Figure 7 shows the performance of VGL and VL in a task of learning 50 trajectories from fixed start points. The VGL learning graph is clearly more smooth, more efficient and more optimal than the VL graph. VGL$\Omega$ and VGL could both manage this task, achieving close to optimal performance each time, but only VGL$\Omega$ could cope with the smaller $c$ values in the range $[0.01, 1)$.

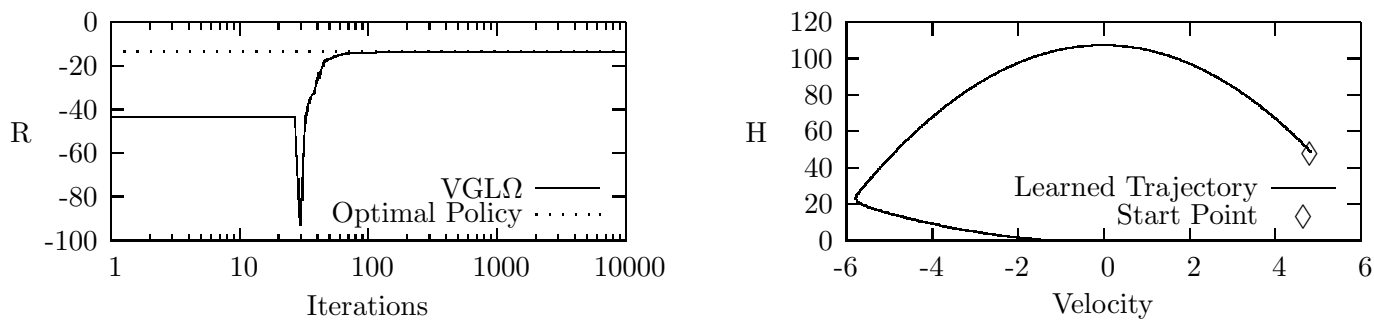

Figure 6: Learning performance for a single trajectory on the Lunar-Lander problem. Parameters: $c = 0.01$, $\Delta t = 0.1$, $\bar{\lambda} = 0$. The algorithm used was VGL$\Omega$. The trajectory produced is very close to optimal (c.f. Fig. 9).

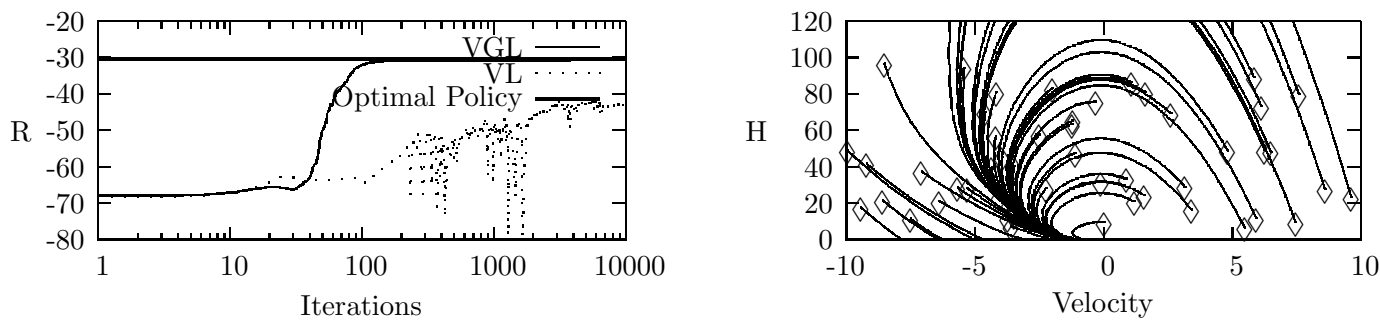

Figure 7: Learning performance for learning 50 trajectories simultaneously on Lunar-Lander problem. Parameters: $c = 1$, $\Delta t = 1$, $\bar{\lambda} = 0$. The algorithms used were VL and VGL. Graphs were averaged over 20 runs. The averaging will have had a smoothing effect on both curves. The right graph shows a set of final trajectories obtained with one of the VGL trials, which are close to optimal.

It was difficult to get VL working well on this problem at all, and the parameters chosen were those that appeared to be most favourable to VL in preliminary experiments. Having such a high $c$ value makes the task much easier, but VL still could not get close to optimal, or stable, trajectories. No stochastic elements were used in either the policy or model. The large number of fixed trajectory start points provided the exploration required by VL. These gave a reasonable sampling of the whole of state space. Preliminary tests with the $\epsilon$-greedy policy did not produce any successful learning. The policy used was the greedy policy described in Appendix E.

We could not get bootstrapping to get to work on this task for either the VL or VGL algorithms. With bootstrapping, the trajectories tended to continually oscillate. It was not clear why this was, but it is consistent with the lack of convergence results for bootstrapping.

It is desirable to have $c$ as small as possible to get fuel-efficient trajectories (see for example Figure 9), but a small $c$ makes the continuous-time DE more stiff. However our experience was that the limiting factor in choosing a small $c$ was not the stiffness of the DE, but the fact that as $c \to 0$, $\frac{\partial R^\pi}{\partial \vec{w}} \to \vec{0}$ which made learning with VGL$\Omega$ difficult. Using RPROP largely remedied this since it copes well with small gradients and can respond quickly when the gradient suddenly changes. We did not need to use a stiff DE solver, and found the Euler method adequate to integrate the equations of Section 2.2.

## 5. Conclusions

This section summarises some of the issues and benefits raised by the VGL approach. Also, the contributions of this paper are highlighted.

- Several VGL algorithms have been stated. These are algorithms VGL($\lambda$), VGL$\Omega$($\lambda$) and VGLRG($\lambda$) (see Table 1), with their continuous time counterparts, and actor-critic learning; all defined for any $0 \le \lambda \le 1$. Results on the Toy Problem and the Lunar-Lander are better than the VL results by several orders of magnitude, in all cases.

- The value-gradient analysis goes a large way to resolving the issue of exploration in value function learning. Local exploration comes for free with VGL. Other than the problem of local versus global optimality, the problems of exploration are resolved, and value function learning is put onto an equal footing with PGL. For example, as discussed in Section 1.5, exploration had previously caused difficulties in Q($\lambda$)-learning and Sarsa($\lambda$).

- In Appendix A, definitions of extremal and optimal trajectories and an optimality proof are given for learning by value-gradients. The proof refers to Pontryagin's Maximal Principle (PMP), but in the case of "bang-bang" control, the conclusion of the proof goes slightly further than is implied solely by PMP.

- The value-gradient analysis provides an overall view that links several different areas within reinforcement learning. For example, the connection between PGL and value function learning is proven in Section 2.1. This provides an explanation of what happens when the "residual gradient terms" are missed off from the weight update equations (i.e. $\frac{\partial E}{\partial \vec{w}} \to \frac{\partial R^\pi}{\partial \vec{w}}$; see Section 2.1), and a tentative justification for the TD($\lambda$) weight update equation (see Section 2.1). Also, the obvious similarity of form between Eq. 15 and Eq. 27 provides connections between PGL and actor-critic architectures, as discussed in Section 3.

- The use of a function approximator has been intrinsic throughout. This followed from the definition of $V$ in Section 1.1. This has led to a robust convergence proof, for a general function approximator and value function, in Section 2.1. The use of a general function approximator is an advancement over simplified function approximators (e.g. linear approximators), or function approximators that require a hand-picked partitioning of state space.

- Most previous studies have separated the value function update from the greedy policy update, but this has been a severe limitation because RL does not work this way; in practice, it is necessary to alternate one with the other, otherwise the RL system would never improve. However no previous convergence proof or divergence example applies to this "whole system". In this paper there has been a tight coupling of the value function to the greedy policy, and this has made it possible to successfully analyse what effect a value function update has to the greedy policy. We have found convergence proofs and divergence examples for the whole system. We do not think it is possible to do this analysis without value-gradients, since the expression for $\frac{\partial \pi}{\partial \vec{w}}$ in Eq. 17 depends on $\frac{\partial G}{\partial \vec{w}}$. Hence value-gradients are *necessary* to understand value-function learning, whether by VL or VGL.

- By considering the "whole system", a divergence example is found for VL (including Sarsa($\lambda$) and TD($\lambda$)) and VGL with $\lambda = 1$, in Section 4.3. This may be a surprising result, since it is generally thought that the case of $\lambda = 1$ is fully understood and always converges, but this is not so for the whole system on a control problem.

- It is proposed in Section 1.4 that the value-gradients approach is an idealised form of VL. We also believe that the approach of VL is not only haphazardly indirect, but also introduces some extra unwanted terms into the weight update equation, as demonstrated in the analysis at the end of Section 4.1.

- The continuous-time policy stated in Section 2.2 (taken from Doya, 2000) is an exact implementation of the greedy policy $\pi(\vec{x}, \vec{w})$ that is smooth with respect to $\vec{x}$ and $\vec{w}$ provided that the function approximator used is smooth. This resolves one of the greatest difficulties of value-function learning, namely that of discontinuous changes to the actions chosen by the greedy policy.

- A new explanation is given about how residual gradient algorithms can get trapped in spurious local minima, as described in Section 2.3. We think this is the main reason why residual gradients often fails to work in practice, even in the case of VGL and deterministic systems. Understanding this will hopefully save other researchers losing too much time exploring this possibility.

It is the opinion of the author that there are several problematic issues about reinforcement learning with an approximated value function, which the algorithm VGL$\Omega(1)$ resolves. These problematic issues are:

- Learning progress is far from monotonic, when measured by either $E$ (Eq. 12) or $R^\pi$, or any other metric currently known. This problem is resolved by the proposed algorithm, when used in conjunction with a policy such as the one in Section 2.2.

- When learning a value function with a function approximator, the objective is to obtain $G_t = G'_t$ for all $t$ along a trajectory. In general, due to the nature of function approximation, this will never be attained exactly. However, even if this is very close to being attained, the value of $R^\pi$ may be far from optimal. In short, minimising $E$ and maximising $R^\pi$ are not same thing unless $E = 0$ can be attained exactly. As demonstrated in the experiment of Section 4.4, and the proofs in Section 2.1, this problem is resolved by the proposed algorithm.

- The success of learning can depend on how state space is scaled. The definition of $\Omega_t$ (Eq. 19) resolves this problem. Other algorithms can become unstable without this.

- Making successful experiments reproducible in VL is very difficult. There are no convergence guarantees, either with or without bootstrapping, and success often depends upon well-chosen parameter choices made by the experimenter. For example, the Lunar-Lander problem in Section 4.5 seems to defeat VL with the given choices of state space scaling and function approximator. With the proposed algorithm, convergence to some fixed point is assured; and so one major element of luck is removed.

All of the proposed algorithms are defined for any $0 \leq \lambda \leq 1$. The results in Section 3.1 and Appendix A are valid proofs for any $\lambda$, but the main convergence result of this paper, Eq. 18, applies only to $\lambda = 1$. Unfortunately divergence examples exist for $\lambda < 1$, as described in Section 4.3.

Also by proving equivalence to policy-learning in the case of $\lambda = 1$, and finding a lack of robustness and divergence examples for $\lambda < 1$, the usefulness of the value-function is somewhat discredited; both for VGL, and its stochastic relative, VL.

## Acknowledgments

I am very grateful to Peter Dayan, Andy Barto, Paul Werbos and Rémi Coulom for their discussions, suggestions and pointers for research on this topic.

## Appendix A. Optimal Trajectories

In this appendix we define locally optimal trajectories and prove that if $G'_t = G_t$ for all $t$ along a greedy trajectory then that trajectory is locally extremal, and in certain situations, locally optimal.

**Locally Optimal Trajectories.** We define a trajectory parametrised by values $\{\vec{x}_0, a_0, a_1, a_2, \ldots\}$ to be locally optimal if $R(\vec{x}_0, a_0, a_1, a_2, \ldots)$ is at a local maximum with respect to the parameters $\{a_0, a_1, a_2, \ldots\}$, subject to the constraints (if present) that $-1 \leq a_t \leq 1$.

**Locally Extremal Trajectories (LET).** We define a trajectory parametrised by values $\{\vec{x}_0, a_0, a_1, a_2, \ldots\}$ to be locally extremal if, for all $t$,

$$\begin{cases} \left(\frac{\partial R}{\partial a}\right)_t = 0 & \text{if } a_t \text{ is not saturated} \\ \left(\frac{\partial R}{\partial a}\right)_t > 0 & \text{if } a_t \text{ is saturated and } a_t = 1 \\ \left(\frac{\partial R}{\partial a}\right)_t < 0 & \text{if } a_t \text{ is saturated and } a_t = -1. \end{cases} \tag{29}$$

In the case that all the actions are unbounded, this criterion for a LET simplifies to that of just requiring $\left(\frac{\partial R}{\partial a}\right)_t = 0$ for all $t$. Having the possibility of bounded actions introduces the extra complication of saturated actions. The second condition in Eq. 29 incorporates the idea that if a saturated action $a_t$ is fully "on", then we would normally like it to be on even more (if that were possible). In fact, in this definition $R$ is locally optimal with respect to any saturated actions. Consequently, if all of the actions are saturated (for example in the case of "bang-bang" control), then this definition of a LET provides a sufficient condition for a locally optimal trajectory.

**Concave Model Functions.** We say a model has concave model functions if all locally extremal trajectories are guaranteed to be locally optimal. In other words, if we define $\nabla_a R$ to be a column vector with $i^{th}$ element equal to $\frac{\partial R}{\partial a_i}$, and $\nabla_a \nabla_a R$ to be the matrix with $(i, j)^{th}$ element equal to $\frac{\partial^2 R}{\partial a_i \partial a_j}$, then the model functions are concave if $\nabla_a R = 0$ implies $\nabla_a \nabla_a R$ is negative definite.

For example, for the two-step Toy Problem with $k = 1$, since $R(x_0, a_0, a_1) = -a_0{}^2 - a_1{}^2 - (x_0 + a_0 + a_1)^2$, we have $\nabla_a \nabla_a R = \begin{pmatrix} -4 & -2 \\ -2 & -4 \end{pmatrix}$, which is constant and negative definite; and so the two-step Toy Problem with $k = 1$ has concave model functions. It can also be shown that the $n$-step Toy Problem with any $k \geq 0$, and any $n \geq 1$, also has concave model functions.

**Constrained Locally Optimal Trajectories.** The previous two optimality criteria were independent of any policy. This weaker definition of optimality is specific to a particular policy, and is defined as follows: A *constrained (with respect to $\vec{w}$) locally optimal trajectory* is a trajectory parametrised by an arbitrary smooth policy function $\pi(\vec{x}_t, \vec{w})$, where $\vec{w}$ is the weight vector of some function approximator, such that $R^\pi(\vec{x}_0, \vec{w})$ is at a local maximum with respect to $\vec{w}$.

If we assume the function $R^\pi(\vec{x}, \vec{w})$ is continuous and smooth everywhere with respect to $\vec{w}$, then this kind of optimality is naturally achieved at any stationary point found by gradient ascent on $R^\pi$ with respect to $\vec{w}$, i.e. by any PGL algorithm.

**Lemma 5** *If $G'_t \equiv G_t$ (for all $t$, and some $\lambda$) along a trajectory found by an arbitrary smooth policy $\pi(\vec{x}, \vec{z})$, then $G'_t \equiv G_t \equiv \left(\frac{\partial R^\pi}{\partial \vec{x}}\right)_t$ for all $t$.*

This lemma is for a general policy, and is required by section 3.1. Here we use the extended definitions of $V'$ and $G'$ that apply to any policy, given in Section 3 and Eq. 26.

First we note that $\left(\frac{\partial R^\pi(\vec{x}, \vec{z})}{\partial \vec{x}}\right)_t \equiv G'_t$ with $\lambda = 1$, since when $\lambda = 1$ any dependency of $G'(\vec{x}_t, \vec{w}, \vec{z})$ on $\vec{w}$ disappears. Also, by Eq. 6 and since $G'_t = G_t$ we get,

$$G'_t = \left(\left(\frac{\partial r}{\partial \vec{x}}\right) + \left(\frac{\partial \pi}{\partial \vec{x}}\right)_t \left(\frac{\partial r}{\partial a}\right)_t\right) + \left(\left(\frac{\partial f}{\partial \vec{x}}\right)_t + \left(\frac{\partial \pi}{\partial \vec{x}}\right)_t \left(\frac{\partial f}{\partial a}\right)_t\right) G'_{t+1}$$

Therefore $G'_t$ is independent of $\lambda$ and therefore $G'_t \equiv G_t \equiv \left(\frac{\partial R^\pi}{\partial \vec{x}}\right)_t$ for all $t$. ∎

**Lemma 6** *If $G'_t \equiv G_t$ (for all $t$, and some $\lambda$) along a greedy trajectory then $G'_t \equiv G_t \equiv \left(\frac{\partial R}{\partial \vec{x}}\right)_t \equiv \left(\frac{\partial R^\pi}{\partial \vec{x}}\right)_t$ for all $t$.*

This is proved by induction. Note that this lemma differs from the previous lemma in that it is specifically for the greedy policy, and the conclusion is stronger. By Eq. 6 and since $G'_t = G_t$ we get,

$$G_t = \left(\frac{\partial \pi}{\partial \vec{x}}\right)_t \left(\frac{\partial Q}{\partial a}\right)_t + \left(\frac{\partial r}{\partial \vec{x}}\right)_t + \left(\frac{\partial f}{\partial \vec{x}}\right)_t G_{t+1}$$

The left term of this sum must be zero since the greedy policy implies either $\left(\frac{\partial \pi}{\partial \vec{x}}\right)_t = \vec{0}$ (in the case that $a_t$ is saturated and $\left(\frac{\partial \pi}{\partial \vec{x}}\right)_t$ exists, by Lemma 2), or $\left(\frac{\partial Q}{\partial a}\right)_t = 0$ (in the case that $a_t$ is not saturated, by Lemma 1). If $\left(\frac{\partial \pi}{\partial \vec{x}}\right)_t$ does not exist then it must be that $\lambda = 0$, since $G'_t$ exists, and when $\lambda = 0$ the definition is $G'_t = \left(\frac{\partial Q}{\partial \vec{x}}\right)_t$. Therefore in all cases,

$$G_t = \left(\frac{\partial r}{\partial \vec{x}}\right)_t + \left(\frac{\partial f}{\partial \vec{x}}\right)_t G_{t+1}$$

Also, differentiating Eq. 1 with respect to $\vec{x}$ gives

$$\left(\frac{\partial R}{\partial \vec{x}}\right)_t = \left(\frac{\partial r}{\partial \vec{x}}\right)_t + \left(\frac{\partial f}{\partial \vec{x}}\right)_t \left(\frac{\partial R}{\partial \vec{x}}\right)_{t+1} \tag{30}$$

So $\left(\frac{\partial R}{\partial \vec{x}}\right)_t$ and $G_t$ have the same recursive definition. Also their values at the final time step $t = F$ are the same, since $\left(\frac{\partial R}{\partial \vec{x}}\right)_F = G_F = \vec{0}$. Therefore, by induction and lemma 5, $G'_t \equiv G_t \equiv \left(\frac{\partial R}{\partial \vec{x}}\right)_t \equiv \left(\frac{\partial R^\pi}{\partial \vec{x}}\right)_t$ for all $t$. ■

**Theorem 7** *Any greedy trajectory satisfying $G'_t = G_t$ (for all $t$) must be locally extremal.*

Proof: Since the greedy policy maximises $Q(\vec{x}_t, a_t, \vec{w})$ with respect to $a_t$ at each time-step $t$, we know at each $t$,

$$\begin{cases} \left(\frac{\partial Q}{\partial a}\right)_t = 0 & \text{if } a_t \text{ is not saturated} \\ \left(\frac{\partial Q}{\partial a}\right)_t > 0 & \text{if } a_t \text{ is saturated and } a_t = 1 \\ \left(\frac{\partial Q}{\partial a}\right)_t < 0 & \text{if } a_t \text{ is saturated and } a_t = -1. \end{cases} \tag{31}$$

These follow from Lemma 1 and the definition of saturated actions. Additionally, by Lemma 6, $G_t \equiv \left(\frac{\partial R}{\partial \vec{x}}\right)_t$ for all $t$. Therefore since,

$$\begin{aligned} \left(\frac{\partial R}{\partial a}\right)_t &= \left(\frac{\partial r}{\partial a}\right)_t + \left(\frac{\partial f}{\partial a}\right)_t \left(\frac{\partial R}{\partial \vec{x}}\right)_{t+1} \\ &= \left(\frac{\partial r}{\partial a}\right)_t + \left(\frac{\partial f}{\partial a}\right)_t G_{t+1} \\ &= \left(\frac{\partial Q}{\partial a}\right)_t \end{aligned}$$

we have $\left(\frac{\partial R}{\partial a}\right)_t \equiv \left(\frac{\partial Q}{\partial a}\right)_t$ for all $t$. Therefore the consequences of the greedy policy (Eq. 31) become equivalent to the sufficient conditions for a LET (Eq. 29), which implies the trajectory is a LET. ■

**Corollary 8** *If, in addition to the conditions of Theorem 7, the model functions are concave, or if all of the actions are saturated (bang-bang control), then the trajectory is locally optimal.*

This follows from the definitions given above of concave model functions and a LET. ∎

**Remark:** In practice we often do not need to worry about the need for concave model functions, since any algorithm that works by gradient ascent on $R^\pi$ will tend to head towards local maxima, not saddle-points or minima. This applies to all VGL algorithms listed in this paper, except for residual-gradients.

**Remark:** We point out that the proof of Theorem 7 could almost be replaced by use of Pontryagin's maximum principle (PMP) (Bronshtein and Semendyayev, 1985), since Eq. 30 implies $\left(\frac{\partial R}{\partial \vec{x}}\right)_t$ is the "costate" (or "adjoint") vector of PMP, and Lemma 6 implies that the greedy policy is equivalent to the maximum condition of PMP. PMP on its own is not sufficient for the optimality proof without use of Lemma 6. Use of PMP would obviate the need for the bespoke definition of a LET that we have used. We did not use PMP because it is only described to be a "necessary" condition for optimality, and the way we have formulated the proof allows us to derive the corollary's extra conclusion for bang-bang control.

## Appendix B. Detailed counterexample of the failure of value-learning without exploration, compared to the impossibility of failure for value-gradient learning.

This section gives a more detailed example than that of Fig. 3, to show why exploration is necessary to VL but not to VGL.

We consider the one-step Top Problem with $k = 1$. For this problem, the optimal policy (Eq. 8) simplifies to

$$\pi^*(x_0) = -x_0/2 \tag{32}$$

Next we define a value function on which a greedy policy can be defined. Let the value function be linear, for simplicity, and be approximated by just two parameters $(w_1, w_2)$, and defined separately for the two time steps.

$$V(x_t, w_1, w_2) = \begin{cases} w_1 + w_2 x_1 & \text{if } t = 1 \\ 0 & \text{if } t = 2 \end{cases} \tag{33}$$

For the final time step, $t = 2$, we have assumed the value function is perfectly known, so that $V_2 \equiv R_2 \equiv 0$. At time step $t = 0$, it is not necessary to define the value function since the greedy policy only looks ahead. Differentiating this value function gives the following value-gradient function:

$$G(x_t, w_1, w_2) = \begin{cases} w_2 & \text{if } t = 1 \\ 0 & \text{if } t = 2 \end{cases}$$

The greedy policy on this value function gives

$$
\begin{aligned}
a_0 &= \pi(x_0, \vec{w}) \\
&= \arg\max_a \left( r(x_0, a, \vec{w}) + V(f(x_0, a), \vec{w}) \right) && \text{by Eqs. 2, 3} \\
&= \arg\max_a \left( -a^2 + w_1 + w_2(x_0 + a) \right) && \text{by Eqs. 33, 7} \\
&= w_2/2
\end{aligned}
\tag{34}
$$

Having defined a value-function and found the greedy policy that acts on it, we next analyse the situations in the VL and VGL cases, each *without* exploration. The value function defined above is used in the following examples.

Note that the conclusions of the following examples cannot be explained by choice of function approximator for $V$. For example Fig. 3 shows a counterexample for a different function approximator, and similar counterexamples for VL can easily be found for any function approximator of a higher degree. A linear function approximator was chosen here since it is the simplest type of approximator that *can* be made to learn an optimal trajectory in this problem, as is illustrated in the VGL example below.

**Value-Learning applied to Toy Problem (without exploration)**: Here the aim is to show that VL, without exploration, can be applied to the one-step Toy Problem (with $k = 1$) and converge to a sub-optimal trajectory.

The target for the value function at $t = 1$ is given by:

$$
\begin{aligned}
V'_1 &= r(x_1, a_1) + \lambda V'_2 + (1 - \lambda)V_2 && \text{by Eq. 4} \\
&= -x_1{}^2 && \text{by Eq. 7, and since } V'_2 = V_2 = 0
\end{aligned}
$$

The value function at $t = 1$ is given by:

$$
V_1 = w_1 + w_2 x_1
$$

A simple counterexample can be chosen to show that if VL is complete (i.e. if $V_t = V'_t$ for all $t > 0$), then the trajectory may not be optimal. If $x_0 = 5$, $w_1 = -25$, $w_2 = 0$ then the greedy policy (Eq. 34) gives $a_0 = w_2/2 = 0$ and thus $x_1 = x_0 = 5$. Therefore $V_1 = V'_1 = -25$, and $V_2 = V'_2 = 0$, and so learning is complete. However the trajectory is not optimal, since the optimal policy (Eq. 32) requires $a_0 = -5/2$. ∎

**Value-Gradient Learning applied to Toy Problem:** The objective of VGL is to make the value-gradients match their target gradients. For the one-step Toy Problem (with $k = 1$), we get:

$$
\begin{aligned}
G'_1 &= \left( \left( \frac{\partial r}{\partial x} \right)_1 + \left( \frac{\partial \pi}{\partial x} \right)_1 \left( \frac{\partial r}{\partial a} \right)_1 \right) \\
&\quad + \left( \left( \frac{\partial f}{\partial x} \right)_1 + \left( \frac{\partial \pi}{\partial x} \right)_1 \left( \frac{\partial f}{\partial a} \right)_1 \right) \left( \lambda G'_2 + (1 - \lambda)G_2 \right) && \text{by Eq. 6} \\
&= (-2x_1 + 0) + (1 + 0) \left( \lambda G'_2 + (1 - \lambda)G_2 \right) && \text{by Eq. 7} \\
&= -2x_1 && \text{since } G'_2 = G_2 = 0
\end{aligned}
$$

The value-gradient at $t = 1$ is given by $G_1 = w_2$.

For these to be equal, i.e. for $G_t = G'_t$ for all $t > 0$, we must have $w_2 = -2x_1$. The greedy policy (Eq. 34) then gives $a_0 = w_2/2 = -x_1 = -(x_0 + a_0) \Rightarrow a_0 = -x_0/2$ which is the same as the optimal policy (Eq. 32). Therefore if the value-gradients are learned, then the trajectory will be optimal. ■

## Appendix C. Equivalence of $V'$ notation in TD($\lambda$)

The formulation of TD($\lambda$) as presented in this paper (Eq. 10) uses the $V'$ notation. This can be proven to be equivalent to the formulation used by Sutton (1988) as follows. Expanding the recursion in Eq. 4 gives $V'_t = \sum_{k \geq t} \lambda^{k-t} (r_k + (1 - \lambda)V_{k+1})$, so Eq. 10 becomes:

$$
\begin{aligned}
\Delta \vec{w} &= \alpha \sum_{t \geq 1} \left( \frac{\partial V}{\partial \vec{w}} \right)_t \left( \sum_{k \geq t} \lambda^{k-t} (r_k + (1 - \lambda)V_{k+1}) - V_t \right) \\
&= \alpha \sum_{t \geq 1} \left( \frac{\partial V}{\partial \vec{w}} \right)_t \left( \sum_{k \geq t} \lambda^{k-t} (r_k + V_{k+1}) - \sum_{k \geq t} \lambda^{k-t} \lambda V_{k+1} - V_t \right) \\
&= \alpha \sum_{t \geq 1} \left( \frac{\partial V}{\partial \vec{w}} \right)_t \left( \sum_{k \geq t} \lambda^{k-t} (r_k + V_{k+1}) - \sum_{k \geq t} \lambda^{k-t} V_k \right) \\
&= \alpha \sum_{k \geq 1} \sum_{t=1}^{k} (r_k + V_{k+1} - V_k) \lambda^{k-t} \left( \frac{\partial V}{\partial \vec{w}} \right)_t \\
&= \alpha \sum_{t \geq 1} (r_t + V_{t+1} - V_t) \sum_{k=1}^{t} \lambda^{t-k} \left( \frac{\partial V}{\partial \vec{w}} \right)_k
\end{aligned}
$$

This last line is a batch-update version of the weight update equation given by Sutton (1988). This validates the use of the notation $V'$.

## Appendix D. Equivalence of Sarsa($\lambda$) to TD($\lambda$) for control problems with a known model

Sarsa($\lambda$) is an algorithm for control problems that learns to approximate the $Q(\vec{x}, a, \vec{w})$ function (Rummery and Niranjan, 1994). It is designed for policies that are dependent on the $Q(\vec{x}, a, \vec{w})$ function (e.g. the greedy policy or $\epsilon$-greedy policy).

The Sarsa($\lambda$) algorithm is defined for trajectories where all actions after the first are found by the given policy; the first action $a_0$ can be arbitrary. The function-approximator update is then:

$$
\Delta \vec{w} = \alpha \sum_{t \geq 0} \left( \frac{\partial Q}{\partial \vec{w}} \right)_t (Q'_t - Q_t) \tag{35}
$$

where $Q'_t = r_t + V'_{t+1}$.

Sarsa($\lambda$) is designed to be able to work in problem domains where the model functions are not known, however we can also apply it to the $Q$ function as defined in Eq. 2 that relies upon our known model functions. This means we can rewrite Eq. 35 in terms of $V(\vec{x}, \vec{w})$ to become exactly the same as Eq. 10. For this reason, we are justified to work with the TD($\lambda$) weight update on control problems using the $\epsilon$-greedy policy in the experiments of this paper.

## Appendix E. One-dimensional Lunar-Lander Problem

In this continuous-time problem, a spacecraft is constrained to move in a vertical line and its objective is to make a fuel-efficient gentle landing. The spacecraft is released from vertically above a landing pad in a uniform gravitational field, and has a single thruster that can produce upward accelerations.

The state vector $\vec{x}$ has three components: height ($h$), velocity ($v$), and fuel remaining ($u$), so that $\vec{x}_t \equiv (h_t, v_t, u_t)^T$. Velocity and height have upwards defined to be positive. The spacecraft can perform upward accelerations $a_t$ with $a_t \in [0, 1]$.

The continuous-time model functions for this problem are:

$$\bar{f}((h, v, u)^T, a) = (v, (a - k_g), -a)^T$$
$$\bar{r}((h, v, u)^T, a) = -(k_f)a + \bar{r}^C(a)$$

$k_g \in (0, 1)$ is a constant giving the acceleration due to gravity; the spacecraft can produce greater acceleration than that due to gravity. $k_f$ is a constant giving fuel penalty. We used $k_g = 0.2$ and $k_f = 2$.

Terminal states are where the spacecraft hits the ground ($h = 0$) or runs out of fuel ($u = 0$). In addition to the continuous-time reward $\bar{r}$ defined above, a final impulse of reward equal to $-v^2 - 2(k_g)h$ is given as soon as the lander reaches a terminal state. The terms in this final reward represent kinetic and potential energy respectively, which means when the spacecraft runs out of fuel, it's as if it crashes to the ground by freefall.

$\bar{r}^C(a) = -\int_{0.5}^{a} g^{-1}(x)dx$ is the action-cost term of the reward function (as described in Section 2.2), where $g(x) = \frac{1}{2}(\tanh(x/c) + 1)$ and therefore $\bar{r}^C(a) = c\left[x \operatorname{arctanh}(1 - 2x) - \frac{1}{2}\ln(1 - x)\right]_{0.5}^{a}$. This means the continuous-time greedy policy is exactly $a_t = g\left(-k_f + \frac{\partial \bar{f}}{\partial a}G_t\right)$ and this ensures $a_t \in (0, 1)$.

The derivatives of these model functions are:

$$\frac{\partial \bar{f}}{\partial \vec{x}} = \begin{pmatrix} 0 & 0 & 0 \\ 1 & 0 & 0 \\ 0 & 0 & 0 \end{pmatrix}, \frac{\partial \bar{r}}{\partial \vec{x}} = \begin{pmatrix} 0 \\ 0 \\ 0 \end{pmatrix}, \frac{\partial \bar{f}}{\partial a} = \begin{pmatrix} 0 & 1 & -1 \end{pmatrix}, \frac{\partial \bar{r}}{\partial a} = -k_f - c \operatorname{arctanh}(2a - 1)$$

### E.1 Discontinuity Corrections for Continuous-Time Formulations (Clipping)

With a continuous-time model and episodic task, care needs to be taken in calculating gradients at any terminal states or points where the model functions are not smooth. Figure 8 illustrates this complication when the spacecraft reaches a terminal state.

This problem means $G'$ changes discontinuously at the boundary of terminal states. Since the learning algorithms only use $G'_t$ for $t < F$, the last gradient we need is the

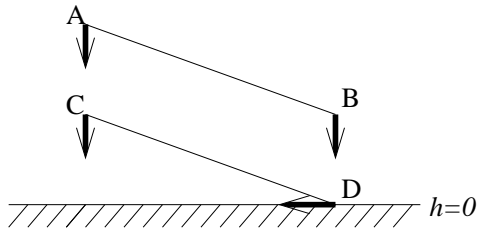

Figure 8: The lines AB and CD are sections of two trajectories that approach a transition to a region of terminal states (the line $h = 0$, in this case). If the end A of AB is moved down then the end B will move down. However if C moves down then D will move left, due to the presence of the barrier. This difference is what we call the problem of Discontinuity Corrections.

limiting one as $t \to F$. This can be calculated by considering the following one-sided limit:

$$
\begin{aligned}
\lim_{\Delta t \to 0^+} \left( \frac{\partial R^\pi}{\partial \vec{x}} \right)_{F - \Delta t} &= \lim_{h \to 0^+} \left( \frac{\partial R^\pi}{\partial \vec{x}} \right)_{F + h/v} \text{ since for small } h, \, \Delta t \approx -h/v \\
&= \lim_{h \to 0^+} \frac{\partial}{\partial \vec{x}_{F + h/v}} \left( ((k_f)a - \bar{r}^C(a)) \frac{h}{v} - \left( v - \frac{(a - k_g)h}{v} \right)^2 \right)_{F + h/v} \\
&= \begin{pmatrix} \frac{(k_f)a_F - \bar{r}^C(a_F)}{v_F} + 2(a_F - k_g) \\ -2v_F \\ 0 \end{pmatrix}
\end{aligned}
$$

where $a_F = \lim_{t \to F^-}(a_t)$.

Similarly it can be shown that, $\lim_{t \to F^-} \left( \frac{\partial R^\pi}{\partial a} \right)_t = 0$ and therefore the boundary condition to use for the target-value gradient is given by

$$
\lim_{t \to F^-} G'_t = \begin{pmatrix} \frac{(k_f)a_F - \bar{r}^C(a_F)}{v_F} + 2(a_F - k_g) \\ -2v_F \\ 0 \end{pmatrix} \tag{36}
$$

This limiting target value-gradient is the one to use instead of the boundary condition given in Eq. 24 or a value-gradient based solely on the final reward impulse. If this issue is ignored then the first component in the above vector would be zero and learning would not find optimal trajectories. A similar correction needs making in the case of the spacecraft running out of fuel.

Also, in the calculation of the trajectory (Eq. 20) by an appropriate numerical method, we think it is best to use clipping in the final time step, so that the spacecraft cannot, for example, finish with $h < 0$. The use of clipping ensures that the total reward is a smooth function of the weights and this should aid methods that work by local exploration.

### E.2 Lunar-Lander Optimal Trajectories

It is useful to know the optimal trajectories, purely for comparison of the learned solutions. Here we only consider optimal trajectories where there is sufficient fuel to land gently. An optimal policy is found using Pontryagin's maximum principle: The adjoint vector $\vec{p}(t)$ satisfies the differential equation (DE)

$$\frac{\partial \vec{p_t}}{\partial t} = -\left(\frac{\partial \bar{r}}{\partial \vec{x}}\right)_t - \left(\frac{\partial \bar{f}}{\partial \vec{x}}\right)_t (\vec{p_t})$$

and the trajectory state evolution equation (i.e. the model functions), and where the action at each instant is found by $a_t = g\left(-k_f + \frac{\partial \bar{f}}{\partial a}p_t\right)$. The trajectory has to be evaluated backwards from the end point at a given $v_F$ and $h = 0$. The boundary condition for this end point is $p_F = \lim_{t \to F^-} G't$ (see Eq. 36) with $a_F = g(-k_f - 2v_F)$.

Solving the adjoint vector DE, and substituting into the expression for $a_t$ gives

$$
\begin{aligned}
\vec{p_t} &= \begin{pmatrix} \bar{p}_F^0 \\ -2v_F + (F-t)\bar{p}_F^0 \\ 0 \end{pmatrix} \text{ with } \bar{p}_F^0 = \frac{(k_f)a_F - \bar{r}^C(a_F)}{v_F} + 2(a_F - k_g) \\
\Rightarrow a_t &= g\left(-k_f - 2v_F + (F-t)\bar{p}_F^0\right)
\end{aligned}
\tag{37}
$$

Numerical integration of the model functions with Eq. 37 gives the optimal trajectory backwards from the given end point. To find which $v_F$ produces the trajectory that passes through a given point $(h_0, v_0)$ is another problem that requires solving numerically. Some optimal trajectories found using this method are shown in Figure 9.

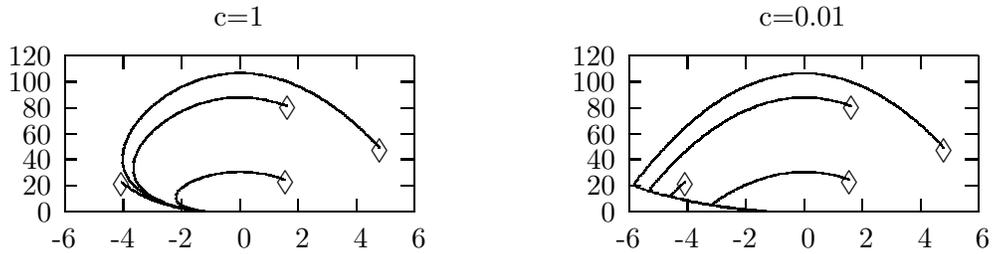

Figure 9: Lunar-Lander optimal trajectories. Shows height (y-axis) vs. velocity (x-axis). Fuel dimension of state-space is omitted. Trajectories start at diamond and finish at $h = 0$. As $c \to 0$, trajectories become more fuel-efficient, and the transition between the freefall phase and braking phase becomes sharper. It is most fuel-efficient to freefall for as long as possible (shown by the upper curved sections) and then to brake as quickly as possible (shown by the lower curved sections).

## Appendix F. Derivation of actor training equation

The actor training equation (Eq. 27) is non-standard. However it can be seen to be consistent with the more standard equations, while automatically including exploration, as follows. Barto et al. (1983) use the TD(0) error signal $\delta_t = (r_t + V_{t+1} - V_t)$ to train the actor, and also specify that some stochastic behaviour is required to force exploration.

When the domain of actions to choose from is continuous, the simplest technique to force exploration is to add a small amount of random noise $n_t$ at time step $t$ to the action chosen (as done by Doya, 2000) giving modified actions $a'_t = a_t + n_t$. The stochastic real-valued (SRV) unit algorithm (Gullapalli, 1990) is used to train the actor while efficiently compensating for the added noise:

$$\Delta \vec{z} = \alpha n_t \left( \frac{\partial \pi}{\partial \vec{z}} \right)_t (r_t + V_{t+1} - V_t)$$

Making a first order Taylor series approximation to the above equation, by expanding the terms $V_{t+1}$ and $r_t$ about the values they would have had if there was no noise, gives

$$\Delta \vec{z} = \alpha n_t \left( \frac{\partial \pi}{\partial \vec{z}} \right)_t \left( r_t + V_{t+1} + n_t \left( \left( \frac{\partial r}{\partial a} \right)_t + \left( \frac{\partial f}{\partial a} \right)_t G_{t+1} \right) - V_t \right)$$

Integrating with respect to $n_t$ to find the mean weight update, and assuming $n_t \in [-\epsilon, \epsilon]$ is a uniformly distributed random variable over a small range centred on zero, gives

$$\langle \Delta \vec{z} \rangle = \int_{-\epsilon}^{\epsilon} \frac{1}{2\epsilon} \Delta \vec{z} dn_t = \alpha \frac{\epsilon^2}{3} \left( \frac{\partial \pi}{\partial \vec{z}} \right)_t \left( \left( \frac{\partial r}{\partial a} \right)_t + \left( \frac{\partial f}{\partial a} \right)_t G_{t+1} \right)$$

which is equivalent to Eq. 27 when summed over $t$. This justifies the use of Eq. 27 and explains how it automatically incorporates exploration. ∎